# Trajectory Tracking Control for Skid-Steering Mobile Robots with Slip and Skid Compensation

Payam Nourizadeh, Fiona J. Stevens-McFadden, Will N Browne

*Abstract*— **Compensating for slip and skid is crucial for mobile robots navigating outdoor terrains. In these challenging environments, slipping and skidding introduce uncertainties into trajectory tracking systems, potentially compromising the safety of the vehicle. Despite research in this field, having a real-world feasible online slip and skid compensation remains challenging due to the complexity of wheel-terrain interaction in outdoor environments. This paper proposes a novel trajectory tracking technique featuring real-world feasible online slip and skid compensation at the vehicle level for skid-steering mobile robots operating outdoors. The approach employs sliding-mode control to design a robust trajectory tracking system, accounting for the inherent uncertainties in this type of robot. To estimate the robot's slipping and undesired skidding and compensate for them in real-time, two previously developed deep learning models are integrated into the control-feedback loop. The main advantages of the proposed technique are that is (1) considers two slip-related parameters for the entire robot, as opposed to the conventional approach involving two slip components for each wheel along with the robot's skidding, and (2) has an online real-world feasible slip and skid compensator, reducing the tracking errors in unforeseen environments. Experimental results demonstrate a significant improvement, enhancing the trajectory tracking system's performance by over 27%.**

## I. INTRODUCTION

WHEELED mobile robots (WMRs) have the capacity to autonomously navigate in on-road and off-road conditions and a wide range of environments such as urban areas, orchards, forests, and planetary exploration. These robots come in various configurations, including independent steering, differential steering, Ackerman steering, and skid-steering, where each category presents its own control issues [1], [2].

Skid-steering mobile robots (SSMRs), unlike other types, lack a dedicated steering mechanism and rely on skidding for steering manoeuvre. This design choice results in lightweight, simplified, and robust robots suitable for off-road terrains and rugged environments. However, the absence of traditional steering mechanisms makes tracking curvilinear trajectories a challenging task for SSMRs [3], [4].

Despite the relatively high traction and robustness, SSMRs still need to deal with terrain-related hazards to be able to operate autonomously in off-road terrains. In off-road environments, the wheel-terrain interaction (WTI) affects the dynamics and controllability of the robot, which cannot be neglected [5]. Consequently, the control system for an autonomous SSMR must account for slipping and skidding when operating outdoors and on uneven terrains. Addressing these issues is vital to minimize tracking errors, prevent immobilization, maintain control, and preserve the robot's mechanical stability.

WTI can cause undesired skidding and slipping for WMRs operating in outdoor environments. Undesired skidding is defined at the vehicle level in the lateral direction and could cause deviation from the desired trajectory [6]. Slippage can be defined at the wheel-level in longitudinal and lateral directions. Hence, a navigation system must determine both longitudinal and lateral slip for each wheel, resulting in two slip parameters per wheel, in addition to accounting for the robot's skidding [7]. Alternatively, the robot's slippage can be assessed at the vehicle-level. In this case, the hypothesis is that the motion control system could only rely on the robot's slipping and undesired skidding (i.e., two parameters in total) for trajectory tracking. This contrasts with the wheel-level perspective, which requires two slip components for each wheel along with the robot's skidding, i.e., commonly nine parameters. Embracing this approach not only reduces the number of necessary slip and skid parameters but also considerably reduces the number of onboard sensors required for their estimation [6], [8].

The main contribution of this paper is the design and real-time implementation of a robust controller for SSMRs with slip and skid compensation at the vehicle-level in outdoor environments and uneven terrains. Initially, the kinematics and dynamics model of an SSMR with slipping and skidding at the vehicle-level is proposed based on the model developed by Pazderski and Kozlowski [9]. Subsequently, a sliding-mode

This work was financially supported by the New Zealand Science for Technological Innovation National Science Challenge (contract CRS-S6-2019); and the Victoria University of Wellington Research Trust.
*(Corresponding author: Payam Nourizadeh).*

Payam Nourizadeh and Fiona J. Stevens-McFadden are with the Robinson Research Institute, Faculty of Engineering, Victoria University of Wellington, Wellington, New Zealand (e-mail: {payam.norizadeh, fiona.stevensmcfadden}@vuw.ac.nz).

Will N Browne is with Faculty of Engineering, School of Electrical Engineering & Robotics, Queensland University of Technology, Brisbane, Australia (e-mail: will.browne@qut.edu.au).

controller is designed based on this dynamics model, aiming to ensure robust performance against model uncertainties during trajectory tracking.

To account for WTI, two deep learning models (i.e., CNN-LSTM-AE and CNN-LSTM algorithms) developed in our prior works [6], [8] are incorporated into the control-feedback loop. These models facilitate estimation for slip and undesired skid at the vehicle-level, enabling the feedback control loop to compensate for these factors in real-time. Notably, these models deliver real-world feasible estimations without relying on prior knowledge of terrain surfaces, utilizing two proprioceptive sensors, i.e., IMU and wheel encoder.

To mitigate the inherent chattering issue associated with the sliding-mode controller, the conventional sign function is replaced with a saturation function, ensuring smoother control actions. Furthermore, this paper investigates and resolves the singularity problem commonly encountered in sliding-mode controllers. It is worth mentioning that singularity points can cause sudden and unpredicted spikes in the control signal, resulting in navigation inaccuracies, non-smooth behaviour, and stability concerns.

The performance of the proposed trajectory tracking system is evaluated using a 4-wheel SSMR (i.e., Pioneer 3-AT) in an outdoor environment. The performance of the proposed controller is also compared with the controller without slip and skid compensation.

The rest of this paper is organized as follows. In Section II, we review related works in trajectory tracking techniques, and slip and skid compensation strategies for mobile robots. Section III presents the modelling of the SSMR with slipping and undesired skidding. Section IV presents the proposed trajectory tracking technique with slip and undesired skid compensation. Section V describes the experimental setup, and the performance of the proposed controllers is evaluated in Section VI. Finally, the conclusion of this paper is presented in Section VII.

## II. Related Works

The problem of modelling and control of SSMRs have been studied in the literature considering the target environment, e.g., indoor or outdoor [4], [10]. In indoor environments, the effect of the WTI is predictable, and as a result under standard operating conditions, the robot's slipping and skidding will be negligible [9], [11]. However, operating the robot in outdoor environments and uneven terrains requires considering the WTI effects on the kinematics and dynamics of the SSMR as well as on the controller design procedure [12]. Note that regardless of the working environment, the SSMR suffers from parametric uncertainties such as the location of the instance centre of rotation.

Due to the nonlinearity of the SSMR, nonlinear controllers have been utilized extensively in the literature [11]. Some researchers have tried to use the linear PID controller for the trajectory tracking of these robots [13], [14]. However, the stability of the closed-loop system might not be guaranteed using that method. The Lyapunov-based controller design technique was utilized as one of the earliest attempts to design a nonlinear trajectory tracking system for SSMR. Other nonlinear controllers were also applied for the trajectory tracking such as nonlinear model predictive control [5], [15], [16] and backstepping techniques [17]. Having a stability proof for the closed-loop system is the main advantage of these techniques [18]. Kozlowski and Pazderski [9] proposed the kinematics and dynamics equations for SSMR without slip and skid consideration, which ensures that their model is valid for indoor environments. They developed a Lyapunov-based trajectory tracking system and validated their technique through simulation and experimental studies. However, the conventional Lyapunov-based controllers do not have the capacity to consider uncertainty in their design procedure and therefore, the trajectory tracking system could be subject to steady-state error due to the parametric uncertainty of this type of robot.

To be able to rectify the parametric uncertainty of SSMRs, Martins et al. [19] integrated an observer into their control-feedback loop. However, observers are dependent on the initial condition and may not provide accurate estimations under different conditions [3]. Model-free techniques were proposed to provide robust trajectory tracking performance including fuzzy logic [20]–[22] and neural networks [23], [24]. Both fuzzy logic and neural network techniques are computationally expensive and might not have stability proof, which could make them difficult to use for real-time experiments.

In outdoor environments, measuring the exact location of the instant centre of rotation of SSMRs is challenging, which causes uncertainty for the control design procedure of these types of robots. Having variable loading is another source of uncertainty that affects the robot's weight and moment of inertia. Therefore, robust controllers such as sliding-mode controllers are common control techniques for this robot as they can guarantee the stability of the closed-loop system considering the uncertainties. Sliding-mode control (SMC) is a well-established robust nonlinear controller that can guarantee the stability of the closed-loop system under disturbances and parametric uncertainty [25]. This controller is designed to force the closed-loop system to a predefined sliding surface (or manifold) considering the uncertainties and varying dynamics of the system. As a result, this controller is robust to system uncertainties and less sensitive to modelling errors. It has been applied to different domains in robotics including holonomic [26], [27] and nonholonomic [3], [28] systems and has shown reliable performance in both simulation and experimental studies. Moreover, this controller is compatible with both single-input-single-output and multi-input-multi-output nonlinear systems, and once the controller is designed, the implementation for experimental studies is simple and computationally efficient in comparison with metaheuristic techniques. These advantages make the sliding-mode controller a suitable candidate for SSMRs' navigation. However, the SMC technique suffers from chattering in the actuators due to having the discontinuous sign function in the control input. To rectify this problem, two approaches mostly have been considered, e.g. (1) using higher-order SMC, or (2) replacing the sign function with continuous alternatives like the saturation function. Note that in the case of replacing the sign function with the continuous one, the stability of the closed-loop system should be investigated. Another inherent challenge with this controller involves the potential singularity issue in some dynamic systems within the equivalent control term, which needs to be investigated during the stability analysis [29]. Matraji et al. [3] implemented a second-order sliding mode controller on an SSMR (Pioneer 3-AT) in indoor environments. They experimentally compared their designed controller with the conventional SMC and showed less chattering in the control inputs. However, their system is an indoor feasible technique only as it did not consider the WTI.

Recent research has addressed the slip and skid problem for motion control of WMRs in different ways such as considering the WTI as an uncertainty [30]–[32], model-based determination of slip and skid [5], [12], [33]–[36], or measuring/estimating [37]–[39] slip and skid to integrate into a control-feedback loop.

The robust control method, like SMC, has the capacity to consider slip and skid as bounded uncertainty. However, as the range of defined uncertainties grows, the steady-state error might increase. The lack of ability to detect high slip/skid conditions (as it is an offline technique) to avoid mentioned terrain-related hazards is another disadvantage of this technique. Therefore, this technique is most suitable for environments with low-slip/skid conditions.

Model-based determination of slip/skid is a classic method that relies on an empirical model of the WTI. The disadvantage of this method is that it requires prior knowledge of soil properties to be able to determine wheel longitudinal and lateral slips, and its accuracy depends significantly on the accuracy of the measured soil properties [33].

Another solution could be measuring or estimating the robot's slip/skid. The slip/skid measurement requires an accurate measurement of the robot's velocity with a sufficient sampling rate for control under different environmental conditions, which in itself is a challenging problem. Alternatively, a slip/skid estimation system can be integrated into the control-feedback loop. This estimator should be real-world feasible, easy to integrate, and be able to operate with a sufficient sampling rate, which has been discussed in the literature [7]. Biswas and Kar [40] proposed a nonlinear observer for slipping and skidding of mobile robots at the vehicle-level based on the kinematics equations of the robot using the Extended Kalman Filter (EKF) technique in indoor environments. They defined the difference between the robot's commanded and observed slip angle as the vehicle skidding and proposed a controller to compensate for the robot's slipping and skidding.

In our previous works [6], [8], we presented novel in-situ slip and undesired skid estimators at the vehicle-level using deep learning and proprioceptive sensors for outdoor environments. In those papers, the deep learning models were trained in outdoor environments for the Pioneer 3-AT robot. Both slip and undesired skid estimators used an IMU and the default wheel encoders, which are low-cost and easy-to-integrate sensors that enabled a real-world feasible estimation system.

In this current paper, we integrate the slip and undesired skid estimators developed in our previous research [6], [8] with a trajectory tracking system to be able to navigate the robot in outdoor environments. The aims of this research are (1) to reformulate the WTI characterization by utilizing two slip and undesired skid parameters at the vehicle-level, as opposed to employing two slip parameters per wheel in addition to the vehicle's skidding, and (2) to design and implement a real-world feasible trajectory tracking system with slip and undesired skid compensation capable of operating in unforeseen outdoor terrains.

## III. Modelling

Figure 1 shows an SSMR with a local coordinate frame $(x_b, y_b, z_b)$ in a 2D plane and $q = [x, y, \theta]^T$ is the location and orientation of the robot's centre of mass (COM) with respect to the global coordinates, $(X, Y, Z)$. $v$ is the velocity of the COM and $\beta$ is the angle of the total velocity of the robot with respect to $x_b$. $v_x$ and $v_y$ are the longitude and lateral velocity components and $\omega$ is the angular velocity of the robot. $2c$ and $r$ are the distance between the rear wheels and the effective radius of the wheels, respectively.

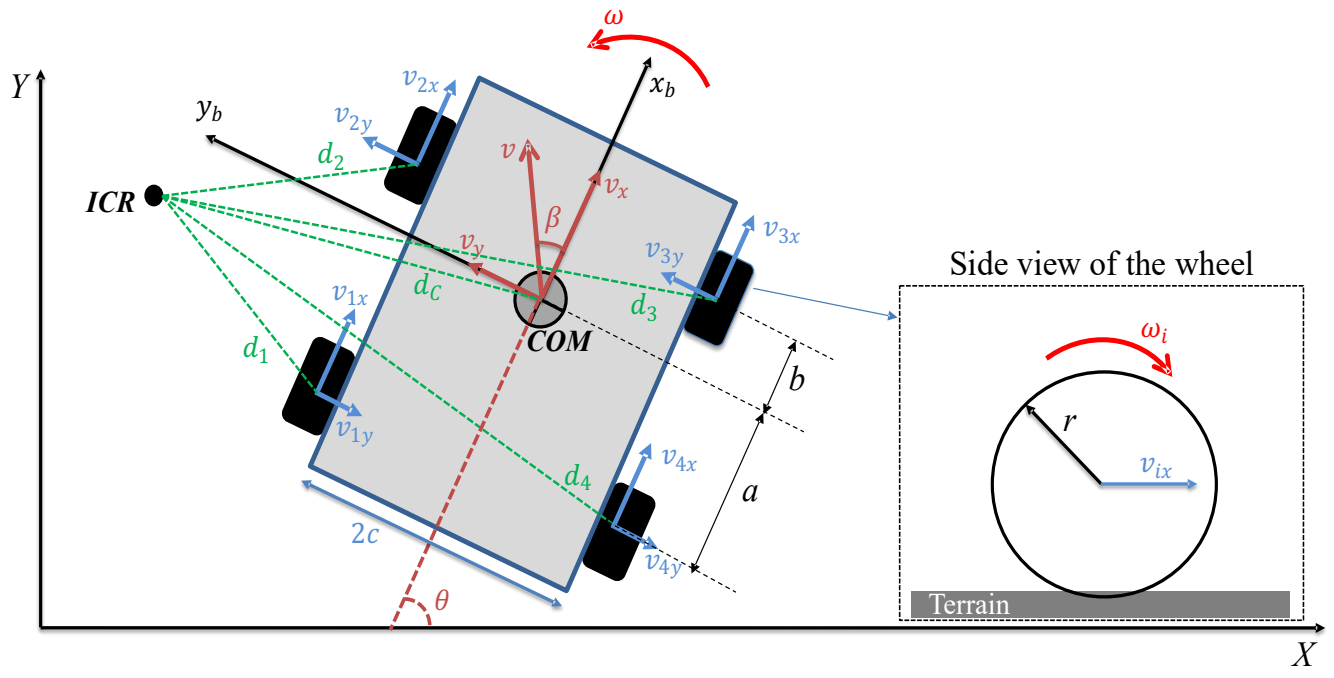

Figure 1. Schematic model of an SSMR with skidding and slipping.

If the robot moves with linear velocity $v = [v_x, v_y, 0]^T$ and angular velocity $\Omega = [0,0,\omega]^T$ expressed in the local frame, then the velocity vector of the robot in the global frame is $\dot{q} = [\dot{x}, \dot{y}, \dot{\theta}]^T$, which based on Figure 1 is

$$\begin{aligned} \dot{x} &= v_x \cos\theta - v_y \sin\theta \\ \dot{y} &= v_x \sin\theta + v_y \cos\theta. \\ \dot{\theta} &= \omega \end{aligned} \quad (1)$$

Eq. (1) provides the relation between the robot's linear and angular velocities and $\dot{q}$ at the COM. The control inputs of the robot are the angular velocity of the wheels $\omega_i$. Therefore, an equation is needed to map the control inputs to the robot's linear and angular velocities. Figure 1 shows wheel interaction with the terrain with $\omega_i$ angular velocity (at the wheel-level). Due to this interaction, the wheel moves in both longitudinal and lateral directions, and the slip at the wheel-level causes a relative velocity at the contact surface. Since this is a rotation with slip, the wheel and terrain interact on a surface instead of a singular point. The linear velocity of the wheels' centre is as follows:

$$v_{ix} = r\omega_i(1 - s_i)$$
$$v_{iy} = v_{ix}\tan\alpha_i \tag{2}$$

In Eq. (2) $s_i$ and $\alpha_i$ are the wheel longitudinal slip ratio and slip angle, respectively.

In Figure 1, *ICR* is the instantaneous centre of rotation and $d_i = [d_{ix}, d_{iy}]^T$ and $d_C = [d_{cx}, d_{cy}]^T$ are the radius vectors and therefore:

$$\begin{bmatrix} d_{1x} \\ d_{2x} \\ d_{1y} \\ d_{3y} \end{bmatrix} = \begin{bmatrix} d_{4x} \\ d_{3x} \\ d_{2y} \\ d_{4y} \end{bmatrix} = \begin{bmatrix} d_{cx} - a \\ d_{cx} + b \\ d_{cy} + c \\ d_{cy} - c \end{bmatrix} \tag{3}$$

and

$$\omega = -\frac{v_{ix}}{d_{iy}} = -\frac{v_x}{d_{cy}} = \frac{v_{iy}}{d_{ix}} = \frac{v_y}{d_{cx}}$$
$$|\omega| = \frac{\|v_i\|}{\|d_i\|} = \frac{\|v\|}{\|d_c\|}, \tag{4}$$

where $\|*\|$ denotes the Euclidian norm, and *a*, *b* and *c* are positive geometrical parameters of the robot. The rigidity constraint can be extracted from Eqs. (3) and (4) as follows:

$$v_L = v_{1x} = v_{2x}, \qquad v_R = v_{3x} = v_{4x}$$
$$v_F = v_{2y} = v_{3y}, \qquad v_B = v_{1y} = v_{4y} \tag{5}$$

In Eq. (5) $v_L$ and $v_R$ are the *x*-component of the left and right wheels' linear velocity respectively, and $v_F$ and $v_B$ are the *y*-component of the front and back wheels' linear velocities respectively. Eq. (5) can be rewritten using Eq. (2).

$$v_L = r\omega_1(1 - s_1) = r\omega_2(1 - s_2)$$
$$v_R = r\omega_3(1 - s_3) = r\omega_4(1 - s_4) \tag{6}$$

The *ICR* can be defined in the robot's local frame as follows:

$$ICR = [x_0, y_0]^T = [-d_{cx}, -d_{cy}]^T \tag{7}$$

Substituting Eq. (7) into (4) gives us:

$$\omega = \frac{v_x}{y_0} = -\frac{v_y}{x_0}. \tag{8}$$

The relationship between the wheels' linear velocity and the linear and angular velocities of the robot at COM can be derived using Eqs. (3)- (8).

$$\begin{bmatrix} v_L \\ v_R \\ v_F \\ v_B \end{bmatrix} = \begin{bmatrix} 1 & -c \\ 1 & c \\ 0 & -x_0 + b \\ 0 & -x_0 - a \end{bmatrix} \begin{bmatrix} v_x \\ \omega \end{bmatrix} \tag{9}$$

The slip ratio $s_i$ is defined for each specific wheel. Assuming all wheels slip independently and the robot moves with the linear $v_x$ and angular $\omega$ velocities, then the longitudinal vehicle slip ratio $s_v$ can be defined at any arbitrary point, e.g., COM, at the vehicle-level as follows:

$$s_v = \frac{v_{Ix} - v_x}{v_{Ix}}, \tag{10}$$

where $v_{Ix}$ denotes the no-slip linear velocity of the robot in the *x* direction, which is:

$$v_{Ix} = r\frac{\omega_L + \omega_R}{2} \tag{11}$$

and therefore:

$$v_x = v_{Ix}(1 - s_v). \tag{12}$$

Note that due to the rigidity constraints (Eq. (5)), the angular velocity of wheels at each side is equal in the no-slip condition.

As also discussed in [6], skidding is the steering mechanism for the SSMR and that means the robot needs skidding to steer. But due to wheel terrain interaction, the robot may experience undesired skidding ($\sigma_v$) that is defined as follows:

$$\sigma_v = v_{Iy} - v_y \tag{13}$$

In the above equation, $v_{Iy}$ denotes the no-slip linear velocity of the robot in the *y* direction. To be able to calculate the undesired skidding, $v_{Iy}$ needs to be determined. For this reason, the nonholonomic constraint of the SSMR needs to be considered.

$$v_{Iy} - x_0\omega_I = 0 \tag{14}$$

In Eq. (14) $\omega_I$ is the no-slip angular velocity of the robot that is defined as follows:

$$\omega_I = r\frac{\omega_R - \omega_L}{2c} \tag{15}$$

Now the linear and angular velocities of the robot with the slip and undesired skid can be written as:

$$\mu = \begin{bmatrix} v_x \\ \omega \end{bmatrix} = \begin{bmatrix} \xi \\ \rho \end{bmatrix},$$

$$\xi = \frac{r}{2}(\omega_R + \omega_L)(1 - s_v) \tag{16}$$

$$\rho = \frac{r}{2c}(\omega_R - \omega_L) + \sigma_v / x_0.$$

The nonholonomic constraint in Eq. (14) can be expressed in the following format as well.

$$A(q)\dot{q} = 0,\ \ A(q) = [-\sin\theta, \cos\theta, x_0] \tag{17}$$

In Eq. (17), $\dot{q}$ is the null space of $A(q)$ and therefore,

$$\dot{q} = R(q)\mu = \begin{bmatrix} \cos\theta & x_0 \sin\theta \\ \sin\theta & -x_0\cos\theta \\ 0 & 1 \end{bmatrix} \begin{bmatrix} v_x \\ \omega \end{bmatrix}. \tag{18}$$

Finally, substituting Eq. (16) in Eq. (18) gives us the kinematics equation of the SSMR with the slip and undesired skidding at the vehicle-level.

$$\begin{bmatrix} \dot{x} \\ \dot{y} \\ \dot{\theta} \end{bmatrix} = R(q) \begin{bmatrix} \xi \\ \rho \end{bmatrix} \tag{19}$$

*Remark 1.* Eq. (16) determines that only two slip parameters ($s_v$ and $\sigma_v$) need to be estimated to be able to control the robot using the vehicle-level slip definition. Whereas in the case of using the wheel-level slip definition, the estimation of the slip ratio for each wheel might be needed [33].

*Remark 2.* It is common for an SSMR to have the same angular speed for each side due to having mechanical coupling, e.g. $\omega_1 = \omega_2 = \omega_L$ and $\omega_3 = \omega_4 = \omega_R$. Then based on Eq. (6) it can be seen that the longitudinal slip for each side of the robot should also be the same, e.g. $s_1 = s_2 = s_L$ and $s_3 = s_4 = s_R$, and therefore:

$$\begin{bmatrix} \omega_L \\ \omega_R \end{bmatrix} = \frac{1}{r} \begin{bmatrix} \dfrac{v_l}{1 - s_L} \\ \dfrac{v_R}{1 - s_R} \end{bmatrix}. \tag{20}$$

*Remark 3.* The robot's undesired skidding was defined in Eq. (13) as the lateral velocity deviation. It also can be defined based on the angle deviation $\delta_v$ as follows:

$$\delta_v = \beta_I - \beta = \tan^{-1}\left(\frac{v_{Iy}}{v_{Ix}}\right) - \tan^{-1}\left(\frac{v_y}{v_x}\right), \tag{21}$$

$$-\pi < \delta_v < \pi$$

In Eq. (21) the $\beta_I$ is the ideal robot's skidding angle (e.g., no undesired skidding). Finally, the input vector $\mu$ in Eq. (16) can be rewritten using Eq. (21).

$$\mu = \begin{bmatrix} v_x \\ \omega \end{bmatrix} = \begin{bmatrix} \xi \\ \Upsilon\xi \end{bmatrix} \tag{22}$$

$$\Upsilon = \frac{\tan\beta}{x_0}$$

It is worth mentioning that, as was discussed in [2], the velocity-based undesired skidding in Eq. (13) offers practical advantages over the angle-based definition in Eq. (21). This is particularly evident in low-speed conditions, where the velocity-based definition exhibits a higher signal-to-noise ratio (SNR).

## IV. Controller Design Procedure

This section presents the design of the sliding mode controller with slip and skid compensation at the vehicle-level.

### *A. Tracking Error Dynamics*

The dynamics equations of the tracking error, *e*, are required to be able to design the controller. Therefore, the tracking errors are defined based on the difference between the desired trajectory and the system states in the global coordinate system as follows:

$$\begin{bmatrix} e_x \\ e_y \\ e_\theta \end{bmatrix} = \begin{bmatrix} x^d - x \\ y^d - y \\ \theta^d - \theta \end{bmatrix} \tag{23}$$

In Eq. (23), $d$ denotes the desired trajectory. To be able to represent the tracking errors in the robot's local frame, the following mapping is applied.

$$\varepsilon = \begin{bmatrix} \varepsilon_1 \\ \varepsilon_2 \\ \varepsilon_3 \end{bmatrix} = \begin{bmatrix} \cos\theta & \sin\theta & 0 \\ -\sin\theta & \cos\theta & 0 \\ 0 & 0 & 1 \end{bmatrix} \begin{bmatrix} e_x \\ e_y \\ e_\theta \end{bmatrix} \tag{24}$$

Taking the time derivative from Eq. (24) gives us the tracking error dynamics.

$$\begin{cases} \dot{\varepsilon}_1 = \frac{d\varepsilon_1}{dt} = -\sin\theta\, \omega e_x + \cos\theta\, \dot{e}_x + \cos\theta\, \omega e_y \\ \qquad\qquad + \sin\theta\, \dot{e}_y \\ \dot{\varepsilon}_2 = \frac{d\varepsilon_2}{dt} = -\cos\theta\, \omega e_x - \sin\theta\, \dot{e}_x - \sin\theta\, \omega e_y \\ \qquad\qquad + \cos\theta\, \dot{e}_y \\ \dot{\varepsilon}_3 = \frac{d\varepsilon_3}{dt} = \dot{\theta}^d - \dot{\theta} \end{cases} \tag{25}$$

Substituting Eq. (24) in Eq. (25) leads to the tracking error dynamics as follows:

$$\begin{cases} \dot{\varepsilon}_1 = \omega\varepsilon_2 + v_x^d \cos\varepsilon_3 + \omega^d x_0 \sin\varepsilon_3 - v_x \\ \dot{\varepsilon}_2 = (x_0 - \varepsilon_1)\omega + v_x^d \sin\varepsilon_3 - \omega^d x_0 \cos\varepsilon_3. \\ \dot{\varepsilon}_3 = \omega^d - \omega \end{cases} \tag{26}$$

*B. Sliding-mode Controller*

In this section, first, the sliding-mode controller is designed based on the tracking error dynamics and the dynamics model of the SSMR. Then the sign function is replaced with the *saturation* function to reduce the controller's chattering and the stability of the controller is demonstrated. Finally, the singularity of the controller is investigated.

The objective of the controller is to regulate the tracking errors $q^d = [\varepsilon_1, \varepsilon_2]$ to force the robot to follow the desired time-varying trajectory considering parametric uncertainty. The controller input is $\mu$ (see Eq. (16)), which requires a multi-input multi-output controller design procedure. Note that due to the robot's physical limitations, the first and second derivatives of the tracking errors are bounded [3].

To design the sliding-mode controller, two sliding manifolds are defined as follows:

$$\begin{cases} s_1 = \lambda_1 \varepsilon_1 + \dot{\varepsilon}_1 \\ s_2 = \lambda_2 \varepsilon_2 + \dot{\varepsilon}_2 \end{cases}, \qquad \lambda_1, \lambda_2 > 0 \tag{27}$$

In Eq. (27), $\lambda_1$ and $\lambda_2$ are positive constants to have stable sliding manifolds. Both sliding manifolds in Eq. (27) are defined based on a proportional-derivative controller, which forces the robot to regulate the tracking error as well as the first derivative of it. Taking the first derivative of the sliding manifolds in Eq. (27) leads to:

$$\begin{aligned} \dot{s}_1 = \frac{ds_1}{dt} = \lambda_1 \dot{\varepsilon}_1 + \ddot{\varepsilon}_1 = \dot{\omega}\varepsilon_2 + \\ \omega[-\omega\varepsilon_1 + v_x^d \sin\varepsilon_3 - \omega^d x_0 \cos\varepsilon_3 + \omega x_0 + \lambda_1 \varepsilon_2] - \\ \lambda_1 v_x - \dot{v}_x + v_x^d[-(\omega^d - \omega)\sin\varepsilon_3 + \lambda_1 \cos\varepsilon_3] + \\ \dot{v}_x^d \cos\varepsilon_3 + \dot{\omega}^d x_0 \sin\varepsilon_3 + \\ \omega^d[x_0(\omega^d - \omega)\cos\varepsilon_3 + \lambda_1 x_0 \sin\varepsilon_3] \end{aligned} \tag{28}$$

and

$$\begin{aligned} \dot{s}_2 = \frac{ds_2}{dt} = \lambda_2 \dot{\varepsilon}_2 + \ddot{\varepsilon}_2 = -\omega\dot{\varepsilon}_1 + \\ (x_0 - \varepsilon_1)\dot{\omega} + \dot{v}_x^d \sin\varepsilon_3 + (\omega^d - \omega) v_x^d \cos\varepsilon_3 - \\ \dot{\omega}^d x_0 \cos\varepsilon_3 + (\omega^d - \omega)\omega^d x_0 \sin\varepsilon_3 + \\ \lambda_2[(x_0 - \varepsilon_1)\omega + v_x^d \sin\varepsilon_3 - \omega^d x_0 \cos\varepsilon_3]. \end{aligned} \tag{29}$$

In Eqs. (28) and (29), we encounter the time derivatives of the controller inputs, i.e., $\dot{v}_x$ and $\dot{\omega}$. To be able to determine these derivatives, the dynamics equation of the robot is required. Since the experimental studies are performed using a commercial SSMR (Pioneer 3-AT), the low-level controller based on the wheels' rotation speeds is already integrated. Therefore, the following dynamics equation is considered [41].

$$\dot{\mu} = \begin{bmatrix} \dot{v}_x \\ \dot{\omega} \end{bmatrix} = \begin{bmatrix} \frac{c_3}{c_1}\omega^2 - \frac{c_4}{c_1} v_x \\ -\frac{c_5}{c_2} v_x \omega - \frac{c_6}{c_2}\omega \end{bmatrix} + \begin{bmatrix} \frac{1}{c_1} & 0 \\ 0 & \frac{1}{c_2} \end{bmatrix} \begin{bmatrix} v_r \\ \omega_r \end{bmatrix} \tag{30}$$

In Eq. (30), $v_r$ and $\omega_r$ are the commanded linear and angular velocities by the low-level controller, and $c_1{:}c_6$ are physical parameters of the robot. Note that as some of these physical parameters might be dependent on hardware and the experimental setup, knowing them precisely could be challenging. Therefore, they are considered with ±25% variation to make the controller robust against the parametric uncertainties.

Substituting Eq. (30) in Eqs. (28) and (29) give us

$$\dot{s}_1 = h_1 + \frac{1}{c_2}\varepsilon_2 \omega_r - \frac{1}{c_1} v_r \tag{31}$$

$$\dot{s}_2 = h_2 + \frac{x_0 - \varepsilon_1}{c_2}\omega_r \tag{32}$$

where,

$$\begin{aligned} h_1 = \left(-\tfrac{c_5}{c_2} v_x \omega - \tfrac{c_6}{c_2}\omega\right)\varepsilon_2 + \omega(-\omega\varepsilon_1 + \\ v_x^d \sin\varepsilon_3 - \omega^d x_0 \cos\varepsilon_3 + \omega x_0 + \lambda_1 \varepsilon_2) - \lambda_1 v_x + \\ \left(-\tfrac{c_3}{c_1}\omega^2 + \tfrac{c_4}{c_1} v_x\right) + v_x^d[-(\omega^d - \omega)\sin\varepsilon_3 + \\ \lambda_1 \cos\varepsilon_3] + \dot{v}_x^d \cos\varepsilon_3 + \dot{\omega}^d x_0 \sin\varepsilon_3 + \omega^d[x_0(\omega^d - \\ \omega)\cos\varepsilon_3 + \lambda_1 x_0 \sin\varepsilon_3], \end{aligned} \tag{33}$$

$$\begin{aligned} h_2 = -\omega(\omega\varepsilon_2 + v_x^d \cos\varepsilon_3 + \omega^d x_0 \sin\varepsilon_3 - v_x) + \\ (x_0 - \varepsilon_1)\left(-\tfrac{c_5}{c_2} v_x \omega - \tfrac{c_6}{c_2}\omega\right) + \dot{v}_x^d \sin\varepsilon_3 + \\ (\omega^d - \omega) v_x^d \cos\varepsilon_3 - \dot{\omega}^d x_0 \cos\varepsilon_3 + (\omega^d - \\ \omega)\omega^d x_0 \sin\varepsilon_3 + \lambda_2[(x_0 - \varepsilon_1)\omega + v_x^d \sin\varepsilon_3 - \\ \omega^d x_0 \cos\varepsilon_3]. \end{aligned} \tag{34}$$

The control inputs are given as follows:

$$v_x^r = \hat{v}_x{}^r + \bar{v}_x{}^r \tag{35}$$

$$\omega^r = \widehat{\omega}^r + \overline{\omega}^r, \tag{36}$$

where $\hat{v}_x{}^r$ and $\widehat{\omega}_r$ are equivalent control terms to stay on the sliding manifolds, and $\bar{v}_r$ and $\overline{\omega}_r$ are the complementary terms to reach the sliding manifolds while dealing with uncertainties.

To determine the equivalent control terms,

$$\dot{s}_2 = 0 \;\rightarrow\; \widehat{\omega}^r = -\frac{c_2 h_2}{\hat{x}_0 - \varepsilon_1} \tag{37}$$

$$\dot{s}_1 = 0 \;\rightarrow \hat{v}_x{}^r = c_1\left(h_1 - \frac{\varepsilon_2 h_2}{\hat{x}_0 - \varepsilon_1}\right), \tag{38}$$

where,

$$\hat{x}_0 = \frac{a + b}{2}, \qquad a \le x_0 \le b, \tag{39}$$

and $a$ and $b$ are physical characteristics of the robot (see Figure 1).

To reach the sliding manifolds in finite time,

$$\dot{s}_i \le -k_i sign(s_i), \qquad i = 1{:}2 \tag{40}$$

where $k_i$ are positive constants. Therefore,

$$\overline{\omega}^r = -\left[\frac{\bar{c}_2}{x_0^{min} - |\varepsilon_1|}\left(-\bar{h}_2 + k_2\right)\right] sign(s_2) \tag{41}$$

$$\bar{v}_x{}^r = -\left[\bar{c}_1\left(\bar{h}_1 + \frac{1}{c_2}\varepsilon_2 \overline{\omega}_r - k_1\right)\right] sign(s_1), \tag{42}$$

where $\bar{*}$ is the maximum amount of that parameter and $x_0^{min}$ is the minimum amount of $x_0$ to guarantee the robustness of the controller.

The design of the sliding-mode controller is therefore completed and Eqs. (35) and (36) are the control inputs. This controller ensures the robust tracking of the desired trajectory with the defined uncertainties. However, due to having the *sign* function in Eqs. (41) and (42), the controller suffers from chattering in actuators [3]. To rectify this problem, the *sign* function is replaced with the *saturation* (*sat*) function in Eqs. (41) and (42) as follows:

$$\bar{\omega}^r = -\left[\frac{\bar{c}_2}{x_0^{min} - |\varepsilon_1|}\left(-\bar{h}_2 + k_2\right)\right] sat\left(\frac{s_2}{\gamma_2}\right) \tag{43}$$

$$\bar{v}_x{}^r = -\left[\bar{c}_1\left(\bar{h}_1 + \frac{1}{c_2}\varepsilon_2\bar{\omega}_r - k_1\right)\right] sat\left(\frac{s_1}{\gamma_1}\right), \tag{44}$$

where $\gamma_1$ and $\gamma_2$ are the positive constants to specify the boundary layer of the *sat* function. The *sat* function addresses the switching issue in the controller. However, the stability of the controller must be thoroughly examined following this modification. For that reason, Eq. (27) is rewritten as follows:

$$\begin{cases}\dot{\varepsilon}_1 = s_1 - \lambda_1\varepsilon_1 \\ \dot{\varepsilon}_2 = s_2 - \lambda_2\varepsilon_2\end{cases}, \tag{45}$$

and coupled with the first derivative of the sliding manifolds gives us:

$$\begin{cases}\dot{\varepsilon}_1 = s_1 - \lambda_1\varepsilon_1 \\ \dot{\varepsilon}_2 = s_2 - \lambda_2\varepsilon_2 \\ \dot{s}_1 = h_1 + \frac{1}{c_1}\varepsilon_2\omega^r - \frac{1}{c_1}v_x^r. \\ \dot{s}_2 = h_2 + \frac{(\hat{x}_0 - \varepsilon_1)}{c_2}\omega^r\end{cases} \tag{46}$$

First, the errors outside of the boundary layer are considered. The Lyapunov function is defined as follows:

$$\begin{aligned} &V_1 = \frac{1}{2}(\varepsilon_1^2 + \varepsilon_2^2), \qquad |s_i| \le \gamma_i, |\varepsilon_i| \ge 2\gamma_i, \\ &\dot{V}_1 = -(\lambda_1\varepsilon_1^2 + \lambda_2\varepsilon_2^2) + \varepsilon_1 s_1 + \varepsilon_2 s_2 \\ &\dot{V}_1 \le -(\lambda_1\varepsilon_1^2 + \lambda_2\varepsilon_2^2) + |\varepsilon_1|\gamma_1 + |\varepsilon_2|\gamma_2 \\ &\dot{V}_1 \le (1-\lambda_1)\varepsilon_1^2 + (1-\lambda_2)\varepsilon_2^2 \\ &\lambda_1, \lambda_2 > 1 \rightarrow \dot{V}_1 \le 0 \end{aligned} \tag{47}$$

Eq. (47) illustrates if $\lambda_1, \lambda_2 > 1$, then the $\dot{V}_1 \le 0$ and as a result, error trajectories from outside the boundary layer get inside the boundary layer.

Now, the stability of the error trajectories needs to be investigated once they get inside the boundary layer. The Lyapunov function is defined as follows:

$$\begin{aligned} &V_2 = \frac{1}{2}(s_2{}^2 + \varepsilon_2^2), \qquad |s_2| \le \gamma_2, |\varepsilon_2| \le 2\gamma_2, \\ &\dot{V}_2 = -\frac{-\bar{h}_2 + \bar{k}_2}{\gamma_2}s_2{}^2 + \varepsilon_2 s_2 - \lambda_2\varepsilon_2^2 \\ &\dot{V}_2 \le -\frac{-\bar{h}_2 + \bar{k}_2}{\gamma_2}s_2{}^2 + |\varepsilon_2||s_2| - \lambda_2\varepsilon_2^2. \end{aligned} \tag{48}$$

In Eq. (48) choosing $\bar{k}_2$ big enough to make $-\bar{h}_2 + \bar{k}_2 = \bar{K}_2 > 0$ gives:

$$\begin{aligned} &\dot{V}_2 \le -\frac{\bar{K}_2}{\gamma_2}s_2{}^2 + |\varepsilon_2||s_2| - \lambda_2\varepsilon_2^2 \\ &\dot{V}_2 \le -[|\varepsilon_2| \quad |s_2|]\begin{bmatrix}\lambda_2 & -\frac{1}{2} \\ -\frac{1}{2} & \frac{\bar{K}_2}{\gamma_2}\end{bmatrix}\begin{bmatrix}|\varepsilon_2| \\ |s_2|\end{bmatrix}. \end{aligned} \tag{49}$$

Choosing $A = \begin{bmatrix}\lambda_2 & -\frac{1}{2} \\ -\frac{1}{2} & \frac{\bar{K}_2}{\gamma_2}\end{bmatrix}$ in Eq. (49) leads to $\dot{V}_2 \le 0$ if $\det(A) \ge 0$. Therefore:

$$\det(A) = \lambda_2\left(\frac{\bar{K}_2}{\gamma_2}\right) - \frac{1}{4} \ge 0 \rightarrow \gamma_2 \le 4\lambda_2\bar{K}_2 \tag{50}$$

Eq. (50) indicates the stability of $s_2$ and $\varepsilon_2$ inside of the boundary layer. The following Lyapunov function investigates the stability of $s_1$ and $\varepsilon_1$.

$$\begin{aligned} &V_3 = \frac{1}{2}(s_1{}^2 + \varepsilon_1^2), \qquad |s_1| \le \gamma_1, |\varepsilon_1| \le 2\gamma_1 \\ &\dot{V}_3 = s_1\left[-\frac{\varepsilon_2\bar{K}_2}{\hat{x}_0 - \varepsilon_1}\left(\frac{s_2}{\gamma_2}\right)\left(1 + \frac{s_1}{\gamma_1}\right) + \bar{K}_1\frac{s_1}{\gamma_1}\right] + \varepsilon_1 s_1 - \lambda_1\varepsilon_1^2, \\ &\dot{V}_3 \le \bar{K}_1\frac{s_1^2}{\gamma_1} - \frac{\varepsilon_2\bar{K}_2}{\hat{x}_0 - \varepsilon_1}\left(\frac{s_2}{\gamma_2}\right)\left(1 + \frac{s_1}{\gamma_1}\right)s_1 + |\varepsilon_1||s_1| - \lambda_1\varepsilon_1^2 \\ &\dot{V}_3 \le \left(\bar{K}_1 - \frac{2\bar{K}_2\gamma_2}{\hat{x}_0 - \varepsilon_1}\right)\left(\frac{s_1^2}{\gamma_1}\right) - \frac{2\bar{K}_2\gamma_2}{\hat{x}_0 - \varepsilon_1}s_1 + |\varepsilon_1||s_1| - \lambda_1\varepsilon_1^2 \end{aligned} \tag{51}$$

In Eq. (51) $\bar{K}_1 = -\bar{h}_1 + \bar{k}_1 > 0$. Substituting Eq. (50) in Eq. (51) gives us:

$$\dot{V}_3 \le \left(\bar{K}_1 + \frac{16\bar{K}_2^2\lambda_2}{|\hat{x}_0| + |\varepsilon_1|} + |\varepsilon_1|\right)\gamma_1 - \lambda_1\varepsilon_1^2$$

If

$$\gamma_1 \le \frac{\lambda_1\varepsilon_1^2}{\bar{K}_1 + \frac{16\bar{K}_2^2\lambda_2}{|\hat{x}_0| + |\varepsilon_1|} + |\varepsilon_1|} \rightarrow \dot{V}_3 \le 0. \tag{52}$$

Eq. (52) ensures the stability of $s_1$ and $\varepsilon_1$ inside the boundary layer.

In conclusion, the above proof shows the stability of the closed-loop system with the *sat* function. Now, the controller inputs in Eqs. (35) and (36) can be updated based on the estimated slip and undesired skid at the vehicle-level provided by the deep learning models as follows [33].

$$v_x^c = \frac{v_x^r}{1 - s_v} \tag{53}$$

$$\omega^c = \omega^r + \sigma_v / x_0 \tag{54}$$

*C. Singularity Analysis*

At this stage, to be able to proceed, a singularity of the controller needs to be investigated, which is caused by the $\hat{x}_0 - \varepsilon_1$ term in the denominator of the $\widehat{\omega}_r$ in Eq. (37). $\hat{x}_0$ is the nominal value of $x_0$ and was chosen in Eq. (39). A singularity happens when $\varepsilon_1 \to \hat{x}_0$, which makes the denominator zero. To avoid the singularity at this specific moment, $\hat{x}_0$ should be chosen in a way to make the nominator of the $\widehat{\omega}_r$ zero as well. If we consider $\hat{x}_0 = 0$, then

$$\begin{aligned} \widehat{\omega}_r &= \frac{c_2 h_2|_{\hat{x}_0=0}}{\varepsilon_1} = \\ &\frac{c_2}{\varepsilon_1}\Big\{-\omega(\omega\varepsilon_2 + v_x^d \cos\varepsilon_3 - v_x) \\ &\quad - \varepsilon_1\left(-\frac{c_5}{c_2} v_x \omega - \frac{c_6}{c_2}\omega\right) \\ &\quad + \dot{v}_x^d \sin\varepsilon_3 + (\omega^d - \omega) v_x^d \cos\varepsilon_3 \\ &\quad - \lambda_2(-\varepsilon_1\omega + v_x^d \sin\varepsilon_3)\Big\} \end{aligned} \tag{55}$$

In the above equation, if $\varepsilon_1 \to 0$, then according to Eq. (24), two situations are expected.

$$\begin{cases} 1.\, \varepsilon_2 \to 0, \quad \varepsilon_3 \to 0 \xrightarrow{yields} \begin{cases} v_x \to v_x^d \\ \omega \to \omega^d \end{cases} \\ 2.\, \varepsilon_2 \to 0, \quad \varepsilon_3 \to \pi \xrightarrow{yields} \begin{cases} v_x \to -v_x^d \\ \omega \to \omega^d \end{cases} \end{cases} \tag{56}$$

In both cases, we have $\widehat{\omega}_r = \frac{0}{0}$ and as a result, L'Hôpital's rule can be applied [42]. For both scenarios

$$\begin{aligned} \widehat{\omega}_r &= \lim_{\varepsilon_1 \to 0} \frac{c_2 \dfrac{dh_2|_{\hat{x}_0=0}}{d\varepsilon_1}}{\dfrac{d\varepsilon_1}{d\varepsilon_1}} \\ &= \lim_{\varepsilon_1 \to 0} \frac{\left(-\frac{c_5}{c_2} v_x\omega - \frac{c_6}{c_2}\omega\right) + \lambda_2\omega}{1} \\ &= \tau \in \mathbb{R}. \end{aligned} \tag{57}$$

Eq. (57) indicates that the singularity for the designed controller can be avoided if zero is chosen for the nominal value of the *x*-component of the robot's *ICR*, e.g., $\hat{x}_0 = 0$.

## V. Slip and Undesired Skid Estimators

This paper utilizes the two previously developed deep learning models [6], [8] to estimate the robot's slipping ($s_v$) and undesired skidding ($\sigma_v$). The structure and details of the recommended models for slip (CNN-LSTM-AE) and undesired skid (CNN-LSTM) estimations are shown in Figure 2 and Table 1. Input sequences for these models are formulated utilizing data derived from the robot's IMU and wheel encoders. The slip estimation model is fed with the robot's commanded linear and angular velocities, roll and pitch angles, as well as angular velocities and linear accelerations in the robot's local frame. Furthermore, the undesired skid estimator receives the same input data along with the change in the yaw angle. The slip and undesired skid estimators consist of a combined total of 83,210 and 129,775 trainable parameters, respectively. Importantly, it should be noted that the training and evaluation of these two models took place in the same location where the trajectory tracking controller was tested in the present study.

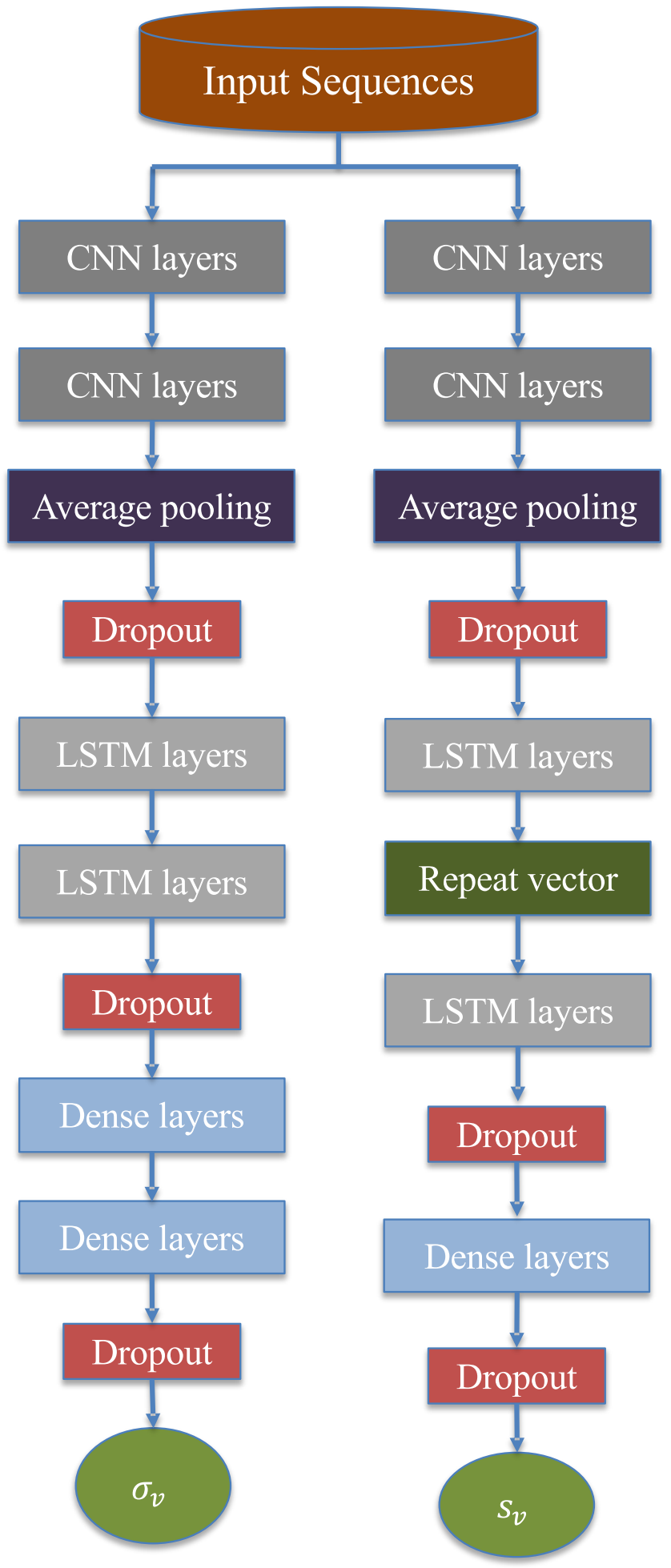

Figure 2. Left: Undesired skid estimation model (CNN-LSTM). Right: Slip estimation model (CNN-LSTM-AE).

Table 1. The hyperparameters of the undesired skid (CNN-LSTM) and slip (CNN-LSTM-AE) estimation models.

| Type | Undesired Skid | Slip |
|---|---|---|
| **Batch size** | 128 | 64 |
| **Conv1D (F, K, S)** | 28, 3, 1 | 67, 5, 1 |
| **Conv1D (F, K, S)** | 32, 3, 1 | 73, 5, 1 |
| **Average Pooling** | 1 | 1 |
| **Dropout** | 0.0 | 0.2 |
| **LSTM unit** | 42 | 44 |
| **LSTM unit** | 121 | 50 |
| **Dropout** | 0.4 | 0.4 |
| **Dense unit** | 131 | 298 |
| **Dense unit** | 112 | - |
| **Dropout** | 0.5 | 0.0 |
| **Dense unit** | 1 | 1 |

## VI. Experimental Setup

The designed sliding-mode controller with slip and skid compensation at the vehicle-level was implemented on a skid-steering Pioneer 3-AT robot as it is a well-established SSMR platform [4]. Testing occurred on outdoor grass terrain characterized by uneven surfaces and varying slope angles, intentionally inducing both slipping and undesired skidding scenarios.

Figure 3 shows the proposed control structure. The Pioneer 3-AT is an all-terrain commercial robot suitable for research purposes. It is equipped with 4 DC motors, high-resolution optical encoders, a microcontroller, and a low-level PID controller. The robot can be controlled using a serial port and a C++ SDK called Advanced Robot Interface for Applications (ARIA), which is developed by the manufacturer [3]. The robot also was equipped with onboard sensors to be able to estimate the slip and skidding as well as for localization purposes. An Xsens IMU (MTi 3-series Development Kit, 100Hz) and an RTK-GPS (U-Blox C94-M8P, 5Hz) were mounted above the middle of the front axle to be able to respectively measure kinematics responses and location/velocity of the robot. A Dell Latitude 5410 was mounted on the robot to collect the sensory data and control the robot in real-time through the robotic operating system (ROS) melodic, Ubuntu 18.04 and Python 3.6. It is worth mentioning that slip and skid estimators require only the IMU and wheel encoder's raw data as the input dataset.

The controller was manually tuned, where Table 2 shows the hyperparameters of the controller. Due to the physical limitation of the robot, two saturation functions were considered in the control-feedback loop having $|v_x^r| \leq 0.5\ m/sec$ and $|\omega^r| \leq 0.3\ rad/sec$.

Table 2. Constant parameters of the controller.

| Parameter | Value |
|---|---|
| $x_0^{max}$ $(cm)$ | 0.15 |
| $x_0^{min}$ $(cm)$ | -0.15 |
| $\lambda_1$ | 1.5 |
| $\lambda_2$ | 1.2 |
| $k_1$ | 5.5 |
| $k_2$ | 2.5 |
| $\gamma_1$ | 0.1 |
| $\gamma_2$ | 0.1 |

In addition to tracking errors, the distance tracking error ($dis$) and the root mean square error (RMSE) were considered to compare the SMC and SMC-SS performance as follows:

$$dis = \sqrt{e_x^2 + e_y^2} \tag{58}$$

$$RMSE = \sqrt{\frac{\sum_{i=1}^{N} e_*^2}{N}}, \tag{59}$$

where $N$ is the number of samples.

Finally, the non-parametric Friedman aligned ranking (FAR) test was utilized to investigate the statistical significance in the performance of the SMC and SMC-SS controllers to check if the SMC and SMC-SS performed similarly. In the case of having the null hypothesis of the FAR test rejected, the post hoc Finner test was applied to determine if there was a significant difference between the two controllers. For both tests, the significance level of 0.05 was considered [43].

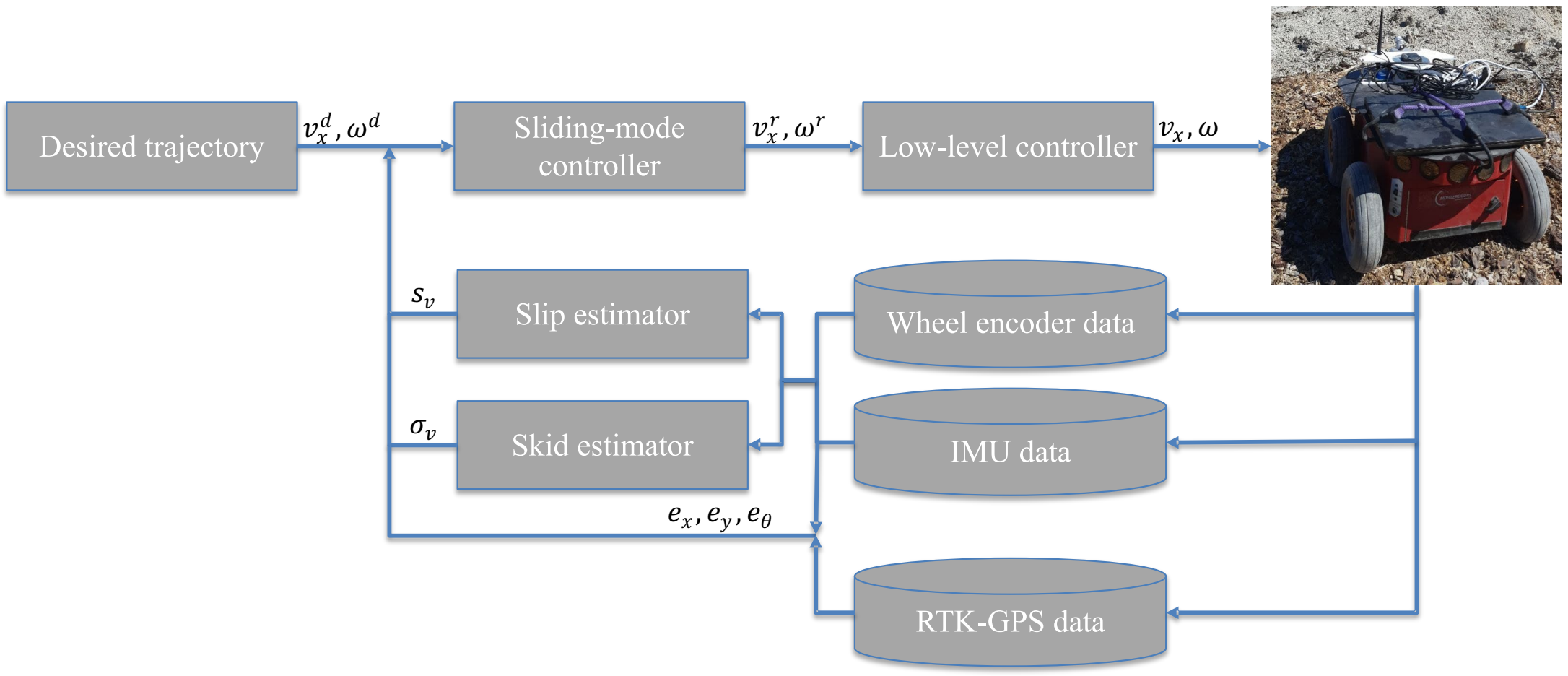

Figure 3. The proposed control block-diagram

## VI. RESULTS

In this section, the performance of the proposed control scheme, incorporating slip and undesired skid compensation for SSMRs in outdoor and uneven terrains, is evaluated. To achieve this objective, we defined three specific manoeuvres for the robot to compare the controller's performance with slip and skid compensation (SMC-SS) and without (SMC). The defined manoeuvres are:

- **Straight-line:** This is a constant trajectory (e.g., $\dot{v}_x^d$ and $\dot{\omega}^d$ are zero) that does not involve the robot's steering mechanism. It tests the controller's ability to track a stable trajectory while maintaining the robot's heading and achieving high SNR for slip estimation and low SNR for undesired skid estimation.
- **Circular:** This trajectory is also constant but requires the robot's steering mechanism. It presents an additional challenge compared to the straight-line manoeuvre as the robot must follow a curvilinear path while compensating for both slip and undesired skidding.
- **Bow-shape:** This manoeuvre is a time-varying trajectory that demands the use of the steering mechanism. It serves to showcase the controller's capabilities in handling a combination of tasks, including following a straight-line, navigating a curvilinear path, and executing stand-still rotations. Additionally, it introduces varying levels of SNR, ranging from low to high, in both slip and undesired skidding scenarios.

These three manoeuvres were chosen based on their varying difficulty levels, introducing complexity for both the SMC and the slip/skid estimators. The experiments aim to demonstrate the controller's competence in diverse operational scenarios, highlighting its capacity for trajectory tracking, stabilization of the robot's heading, and effective compensation for slip and undesired skidding across different challenging terrains and manoeuvres.

Note that for all three manoeuvres, the robot started from the same location with the following initial errors to assess the controllers' performance across both transient and steady-state phases ($e_x$ and $e_y$ are in *m*, and $e_\theta$ in *rad*).

$$\begin{bmatrix} e_x \\ e_y \\ e_\theta \end{bmatrix} = \begin{bmatrix} 0.3 \\ 0.1 \\ 0.0 \end{bmatrix} \tag{60}$$

### *A. Straight-Line Trajectory*

For this experiment, the robot started from the initial location with the initial errors of Eq. (60). The desired trajectory was defined as a straight-line with constant linear and angular velocities as follows:

$$v_x^d = 0.3\, m/sec, \qquad \dot{v}_x^d = 0\, m/sec^2$$

$$\omega^d = 0\, rad/sec, \qquad \dot{\omega}^d = 0\, rad/sec^2$$

Note that the desired trajectory is generated based on a virtual robot operating in an ideal condition without slipping and skidding. The Pioneer robot was driven using the SMC and then SMC-SS controllers.

The robot was driven three times for each controller and the results are in Table 3. This table shows that the SMC-SS controller consistently improved the tracking performance of the robot in following the defined straight-line for all three experiments. In this section, the results of the first experiment are visualized and the results of the second experiment are in the Appendix. The average performance of the SMC and SMC-SS controllers is given in Table 3.

Table 3. Performance of SMC and SMC-SS controllers tracking the straight-line trajectory under uneven grass terrain conditions. M: mean of, *dis* in cm, and $e_\theta$ in degree

| **Trial** | **SMC, SMC-SS** | | | |
|---|---|---|---|---|
| | **M *dis*** | **RMS *dis*** | **M $\lvert e_\theta \rvert$** | **RMS $e_\theta$** |
| **1** | 14.83, 10.45 | 27.04, 19.37 | 3.38, 1.82 | 4.54, 2.31 |
| **2** | 15.50, 11.20 | 27.12, 21.93 | 3.89, 2.00 | 5.0, 2.77 |
| **3** | 16.33, 11.99 | 26.51, 21.79 | 3.69, 1.71 | 4.5, 2.34 |
| **Average** | 15.55, 11.21 | 26.89, 21.03 | 3.65, 1.84 | 4.7, 2.47 |

Figure 4 to Figure 6 and Table 3 show the experimental results of the proposed SMC and SMC-SS controllers following a straight-line versus time. Figure 4-top shows that with each controller the robot starts from the origin and tries to regulate the initial error and then stay on the desired trajectory. Figure 4-bottom and Table 3 show that using the SMC-SS controller, the robot completes the manoeuvre with less tracking error and experiences on average 27.91% and 21.79% improvement in mean and RMS of distance error, which would be an important consideration when driving in a narrow passage, for example, an orchard.

According to Figure 5 and Figure 6-top and middle, the robot with the SMC-SS controller converges faster to the desired trajectory with less overshoot in both *x* and *y* directions and stays on the trajectory with less variation around the *y*-axis. The compensation for vehicle slipping ($s_v$) yielded a faster convergence time and less tracking error in the *x* direction.

Due to the unevenness of the grassy terrain and the WTI, the robot experiences fluctuation in the heading angle with both controllers (Figure 5 and Figure 6-bottom). The unevenness of the terrain directly affects the orientation of the robot and causes undesired skidding for the robot. The other reason for the robot's undesired skidding during this manoeuvre could be because of having different wheel longitudinal slippage on each side of the robot. The objective for the robot was to follow a straight-line trajectory, which means having the same linear speed for wheels on both sides of the vehicle. However, different longitudinal slips for each side of the robot could cause different relative linear speeds between the left and right wheels, and as a result, generate undesired skidding and cause fluctuation in the heading angle for the robot. Although both controllers aimed to regulate the $q^d = [\varepsilon_1, \varepsilon_2]$, the SMC-SS steers the robot with 47.56% less heading angle fluctuation in the RMS of $e_\theta$ due to its consideration of the vehicle slipping and undesired skidding (Table 3).

Figure 7 shows the robot's actual undesired skidding and slipping at the vehicle-level during the straight-line manoeuvre. Figure 7-left explains $e_y$ and $e_\theta$ in Figure 6 for this manoeuvre. For example, the jump of $e_y$ and $e_\theta$ with the SMC controller between 15 and 20 seconds are related to the undesired skidding at this moment for the robot. Figure 7-right shows that the robot experiences respectively about 60% and 80% of slip at the beginning of the manoeuvre with SMC-SS and SMC controllers, where more measurement and estimation error might be expected due to relatively lower SNR at such moments [6], [8]. Then the robots' slipping reduces to about 20% for the remainder of the manoeuvre with both controllers. It is important to emphasize that the aim of the controller was not to reduce the amount of the robot's slip and undesired skid but rather to accurately follow the given trajectory. Consequently, the assessment of the controllers' performance does not centre around the quantification of these two parameters.

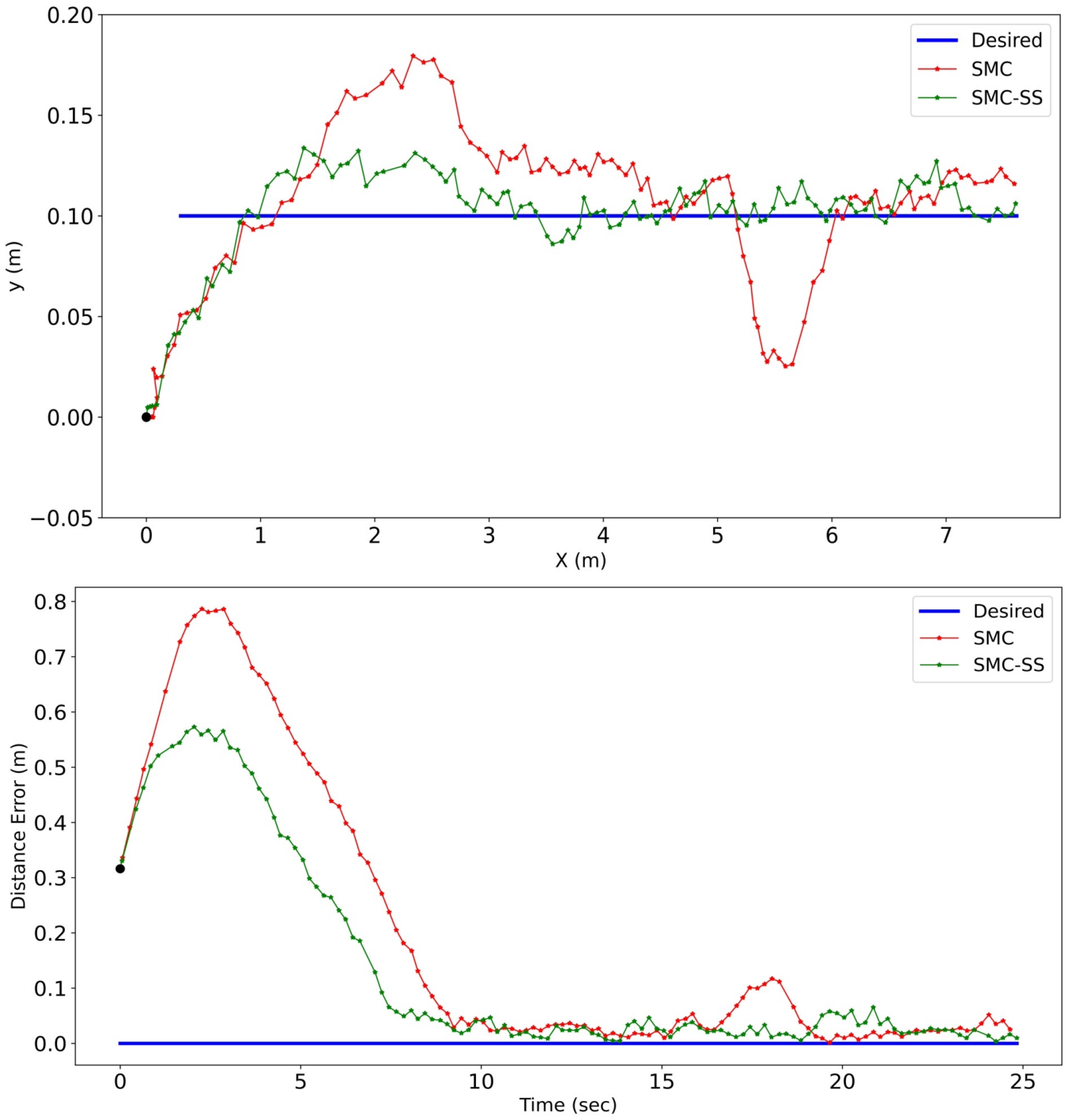

Figure 4. Experimental results of the SMC and SMC-SS controllers following the straight-line trajectory; (top) trajectory, (bottom) distance error.

*B. Circular Trajectory*

The performance of the SMC and SMC-SS was evaluated following a curvilinear trajectory. This is a challenging trajectory for the SSMR due to the skid-based steering mechanism of this robot. The following desired linear and angular velocities for the robot were considered.

$$v_x^d = 0.2\ m/sec, \qquad \dot{v}_x^d = 0\ m/sec^2$$
$$\omega^d = 0.2\ rad/sec, \qquad \dot{\omega}^d = 0\ rad/sec^2$$

The robot was driven three times to follow the defined circular trajectory using the SMC and SMC-SS controllers (Table 4). Similar to the previous section, the following figures show the first experiment and the results of the second experiment are in the Appendix for comparison. The average performance of each controller is given in Table 4.

Table 4. Performance of SMC and SMC-SS controllers tracking the circular trajectory under uneven grass terrain conditions. M: mean of, *dis* in cm, and $e_\theta$ in degree

| **Trial** | **SMC, SMC-SS** | | | |
|---|---|---|---|---|
| | **M *dis*** | **RMS *dis*** | **M $\lvert e_\theta \rvert$** | **RMS $e_\theta$** |
| **1** | 11.38, 6.52 | 17.32, 10.76 | 13.46, 11.27 | 15.51, 12.42 |
| **2** | 12.34, 7.28 | 20.72, 12.59 | 14.12, 12.20 | 14.79, 13.63 |
| **3** | 13.01, 8.86 | 19.95, 14.94 | 13.37, 12.78 | 13.90, 13.99 |
| **Average** | 12.24, 7.55 | 19.33, 12.76 | 13.65, 12.08 | 14.73, 13.34 |

Figure 8 to Figure 10, Table 4 show the performance of the designed controllers on the circular trajectory. Figure 8-top shows that the robot starts with the initial error and converges to the circular trajectory whilst trying to regulate the steady-state error. Figure 8-bottom shows the distance error of both controllers. In contrast to the straight-line trajectory, both controllers perform similarly at the beginning of the manoeuvre and experience almost similar overshoot and convergence time. However, according to Table 4, the SMC-SS reduces the RMS of distance error by 33.99% and the robot on average experiences less distance error with this controller by 38.32%.

As shown in Figure 9 and Figure 10-top and middle, both controllers performed similarly at the beginning of the trajectory in the *x* and *y* directions, but the SMC-SS provides better performance in regulating the steady-state error. Note that during this manoeuvre, the robot needs to constantly change its orientation to be able to stay on the desired path. Therefore, the undesired skidding has an important influence during the entire manoeuvre, and the ability of the SMC-SS to compensate for it led to a better performance in regulating the $e_x$ and $e_y$. That ability also helped the robot to achieve less mean and RMS of $e_\theta$ by 11.50% and 9.44%, respectively (see Figure 9 and Figure 10-bottom, and Table 4).

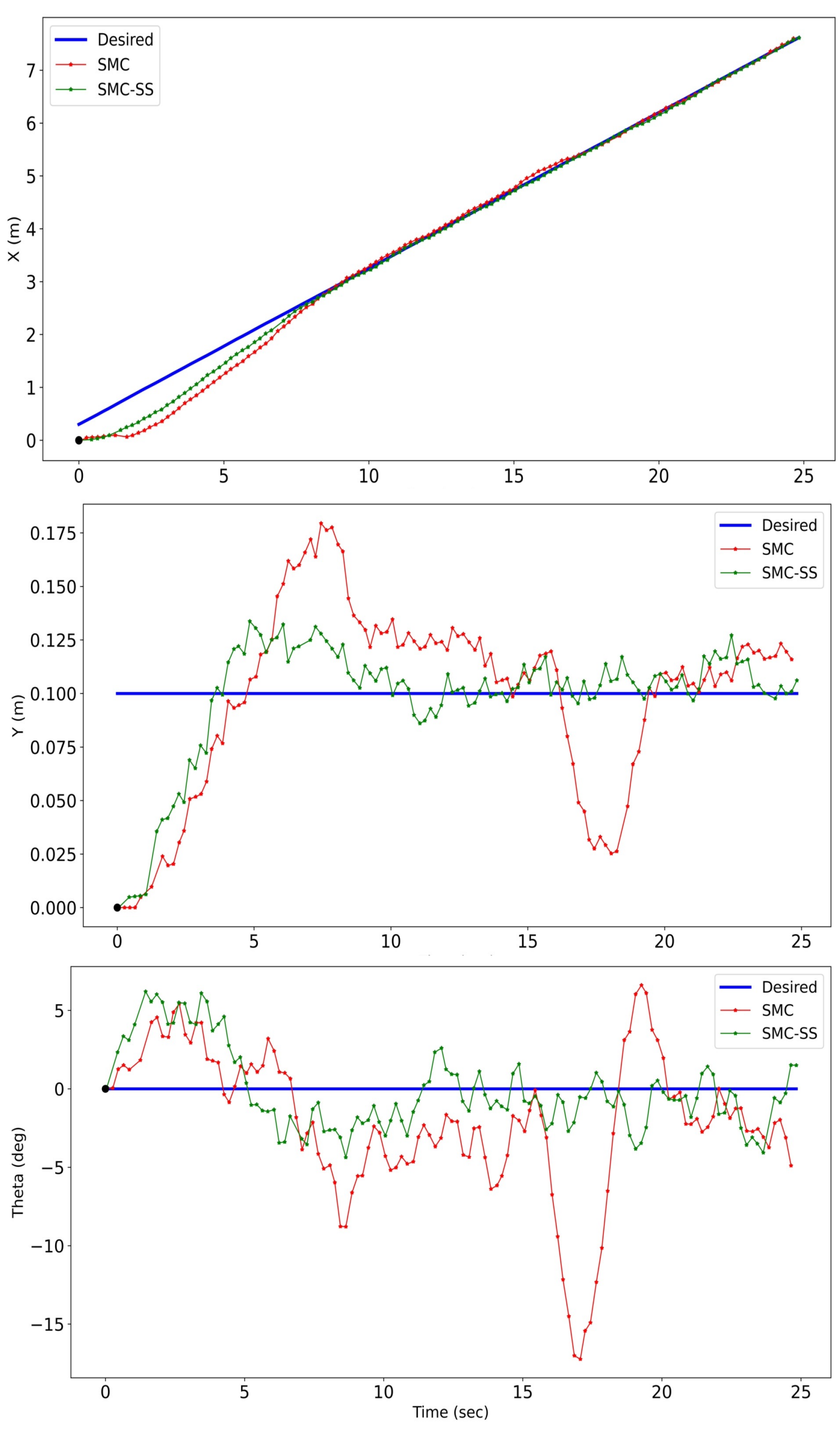

Figure 5. Trajectories of the SMC and SMC-SS controllers following the straight-line trajectory; (top) $x$, (middle) $y$, (bottom) $\theta$.

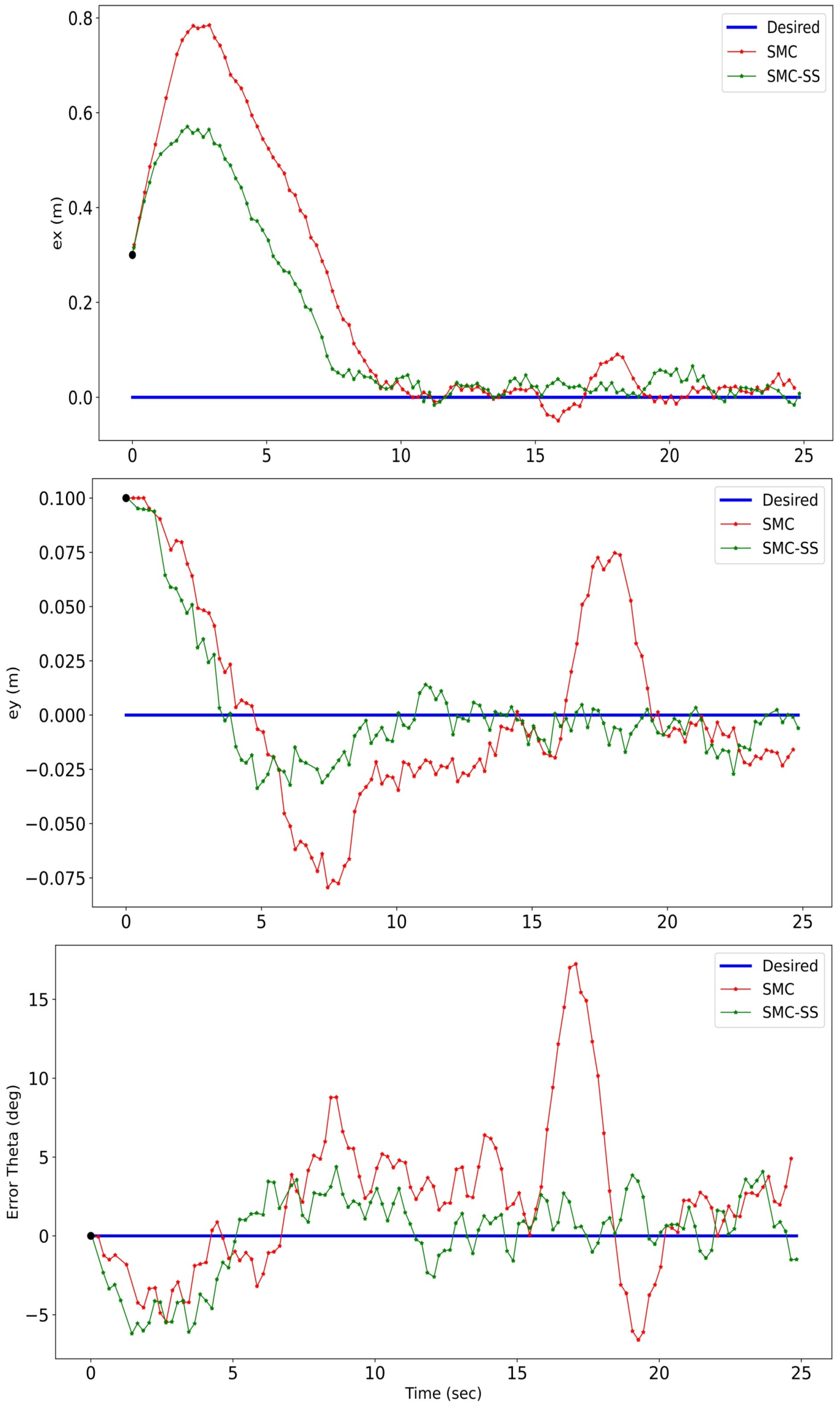

Figure 6. Tracking errors of the SMC and SMC-SS controllers following the straight-line trajectory; (top) $e_x$, (middle) $e_y$, (bottom) $e_\theta$.

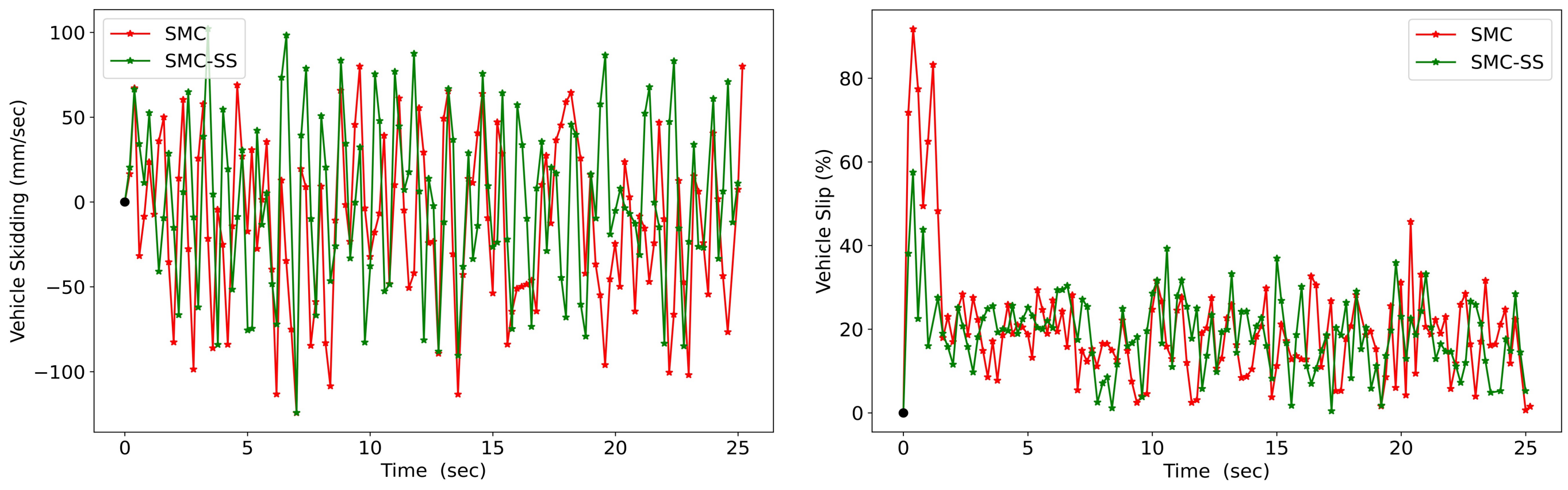

Figure 7. The robot's actual undesired skidding (left) and slipping (right) following the straight-line trajectory.

Figure 11 shows the robot's undesired skidding and slipping at the vehicle-level on the circular trajectory. Figure 11-left shows that the robot experiences a high amount of undesired skidding with both controllers between 20 to 25 seconds, but the compensation ability of the SMC-SS helps the robot to pass this moment with relatively less $e_y$ and $e_\theta$. In addition, the high slippage between 18 to 20 seconds with the SMC controller causes a tracking error for the robot in the *x*-direction (Figure 11-right), which takes time for the controller to reduce the tracking error due to not having a real-time compensation system for it (Figure 10-top).

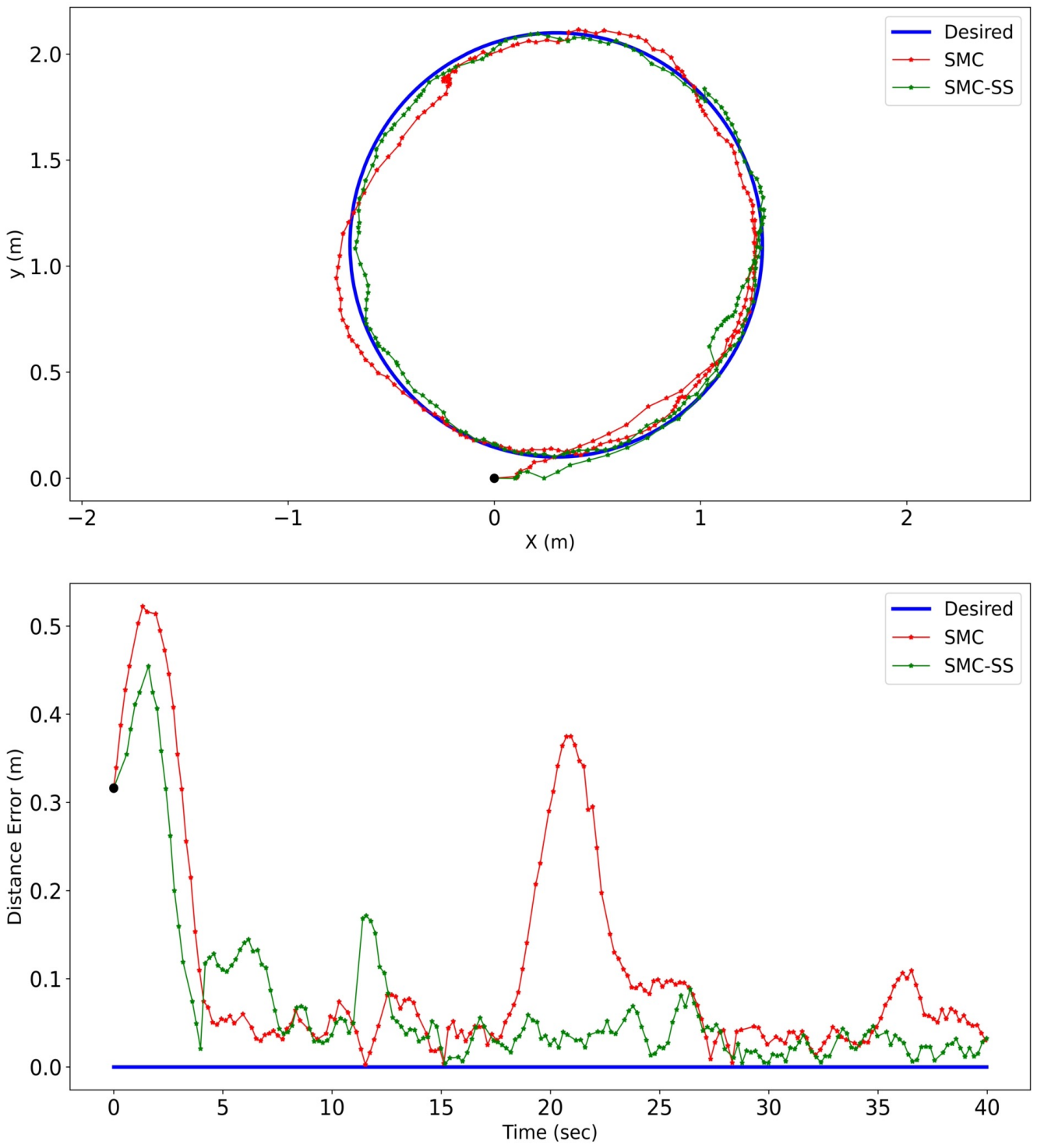

Figure 8. Experimental results of the SMC and SMC-SS controllers following the circular trajectory; (top) trajectory, (bottom) distance error.

*C. Bow-Shape Trajectory*

The last test was conducted to evaluate the performance of the SMC-SS in comparison with the SMC for following a more complicated trajectory. The bow-shape trajectory, similar to the circular one, requires the steering mechanism of the robot during the manoeuvre. Moreover, this trajectory has cusp points where the robot needs to change its orientation sharply. At the cusp points, the robot needs a pure rotation, which is one of the abilities of the SSMR. The desired trajectory for the robot is defined as follows:

$$v_x^d = 0.2\sin(0.1t)\ m/sec,$$

$$\dot{v}_x^d = 0.02\cos(0.1t)\ m/sec^2$$

$$\omega^d = 0.2\cos(0.1t)\, rad/sec,$$

$$\dot{\omega}^d = -0.02\sin(0.1t)\ rad/sec^2$$

For this manoeuvre, two experiments were performed with both controllers and the overall performance of the controllers is given in Table 5. The results of the first and second experiments are visualized in this section and the Appendix, respectively.

Table 5. Performance of SMC and SMC-SS controllers tracking the bow-shape trajectory under uneven grass terrain conditions. M: mean of, *dis* in cm, and $e_\theta$ in degree

| Trial | SMC, SMC-SS | | | |
|---|---|---|---|---|
| | **M *dis*** | **RMS *dis*** | **M $\lvert e_\theta \rvert$** | **RMS $e_\theta$** |
| **1** | 9.16, 6.83 | 12.96, 9.18 | 20.07, 18.94 | 36.27, 32.42 |
| **2** | 10.32, 7.31 | 14.66, 11.16 | 21.51, 18.10 | 35.13, 31.91 |
| **Average** | 9.74, 7.07 | 13.81, 10.17 | 20.79, 18.52 | 35.70, 32.17 |

The experimental results of the robot following the bow-shape trajectory are presented in Figure 12 to Figure 14, Table 6, and Table 7. During this manoeuvre, the robot needs to constantly change its orientation as well as pass two sharp rotations at cusp points. Figure 12-top shows the trajectory of the robot with both controllers. This figure shows that the robot handles the two sharp rotations and stays on the desired trajectory with less error using the SMC-SS controller. On the other hand, the robot experiences a noticeable overshoot after passing each cusp point with the SMC controller. Figure 12-bottom shows that the robot converges slightly faster with the SMC-SS controller and according to Table 7, it experiences on average less mean and RMS of distance error by 27.41% and 26.36% respectively.

Figure 13 and Figure 14-top and middle show the *x* and *y*-component of the robot's trajectory and their equivalent tracking errors during this manoeuvre. In these figures, the first and second cusp points respectively happen around 40 and 80 seconds and the tracking error before and after these points illustrates the performance improvement by the SMC-SS controller.

Figure 15 shows the robot's actual undesired skidding and slipping at the vehicle-level following the bow-shape trajectory. Note that in this case, the longitudinal and lateral movements of the robot in the global coordinate system are in the *y* and *x*-direction, respectively. Therefore, the undesired skidding and slipping of the robot concerns mostly $e_x$ and $e_y$ respectively. Figure 15-left shows that the robot experiences jumps in undesired skidding with the SMC controller at around 50, 60, and 90 seconds, which equivalently causes high tracking errors in the *x*-direction with this controller (Figure 14-top). Figure 15-right shows two continuous high slip conditions ($s_v \geq 70\%$) with both controllers around 40 and 80 seconds as the robot passes cusp points, which the robot experiences low-speed high slippage. On the other hand, this figure shows high slippage with the SMC-SS controller after the first cusp point at around 50 seconds.

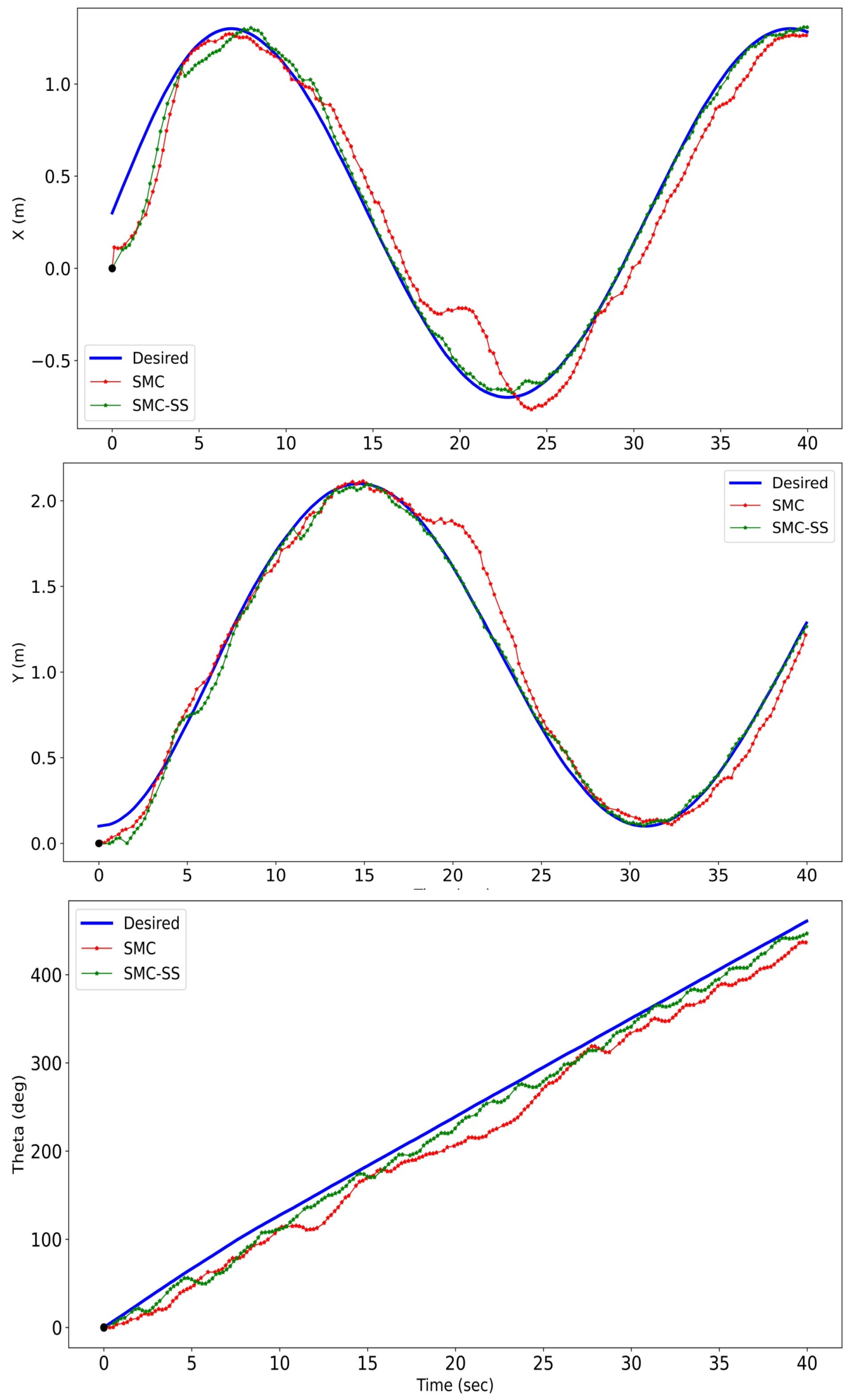

Figure 9. Trajectories of the SMC and SMC-SS controllers following the circular trajectory; (top) $x$, (middle) $y$, (bottom) $\theta$.

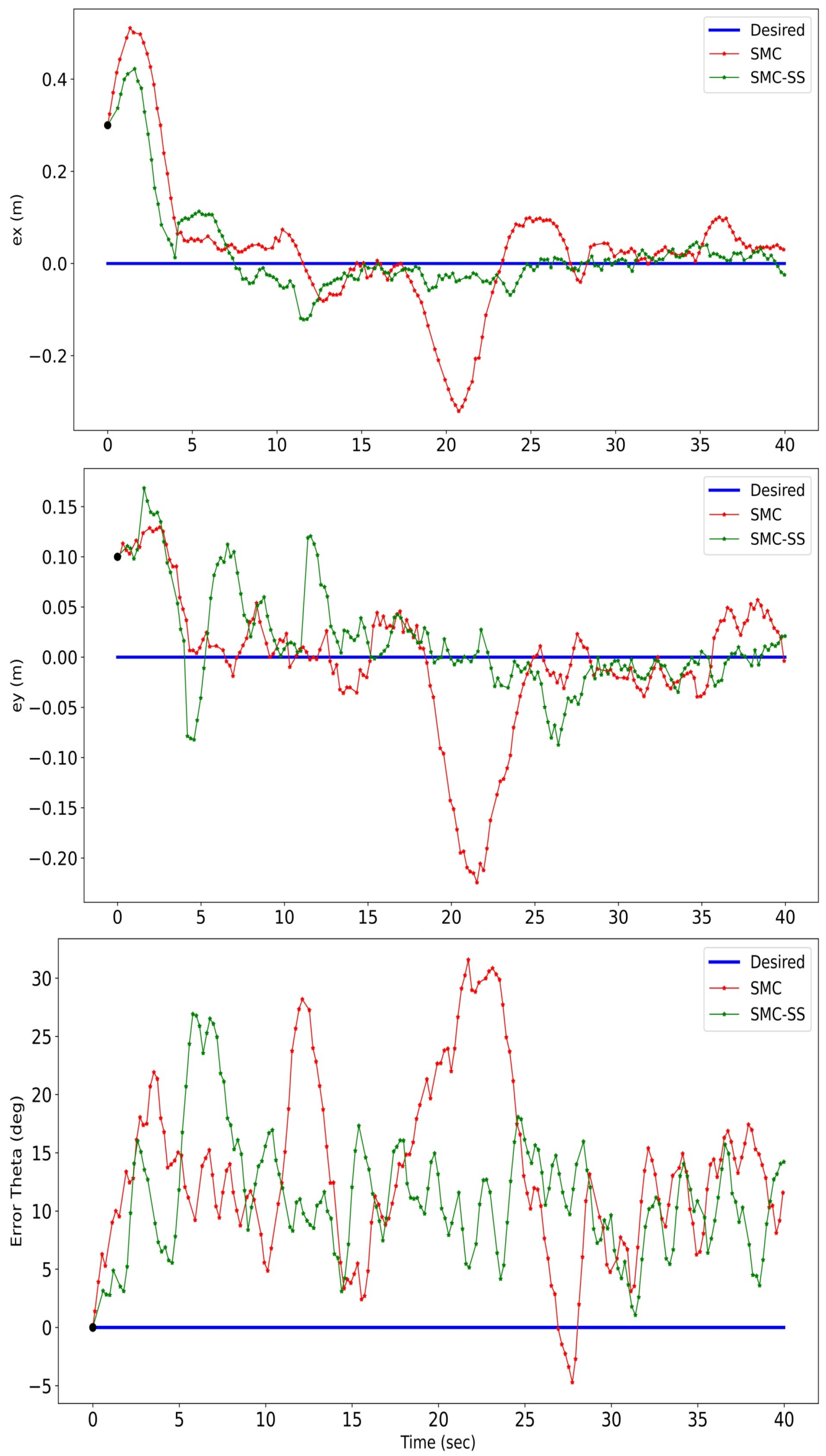

Figure 10. Tracking errors of the SMC and SMC-SS controllers following the circular trajectory; (top) $e_x$, (middle) $e_y$, (bottom) $e_\theta$.

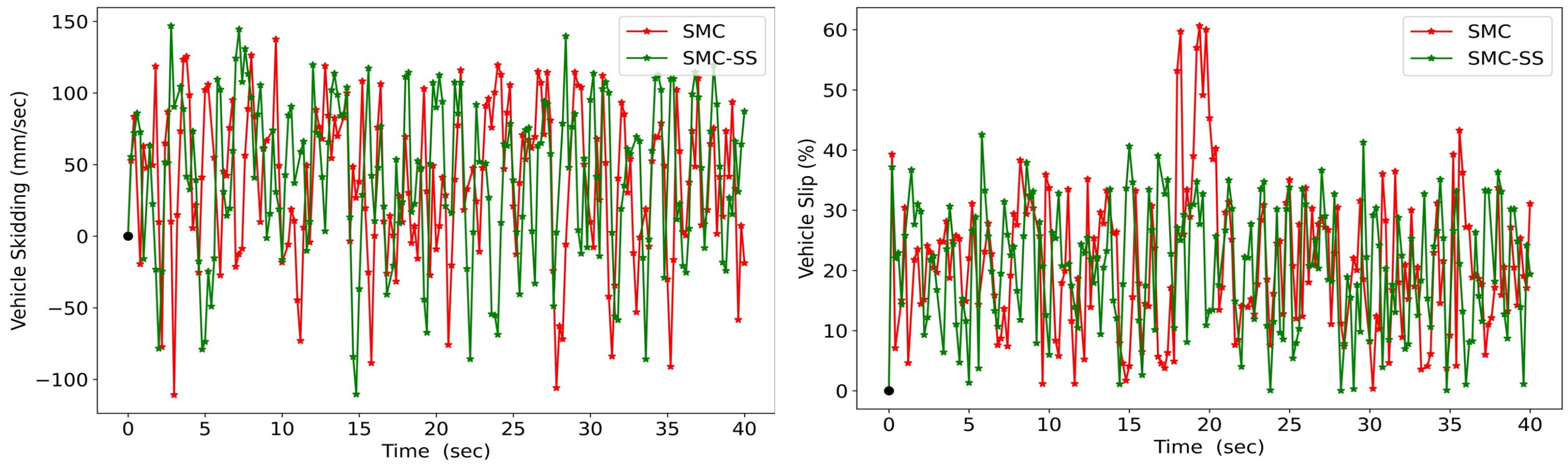

Figure 11. The robot's actual undesired skidding (left) and slipping (right) following the circular trajectory.

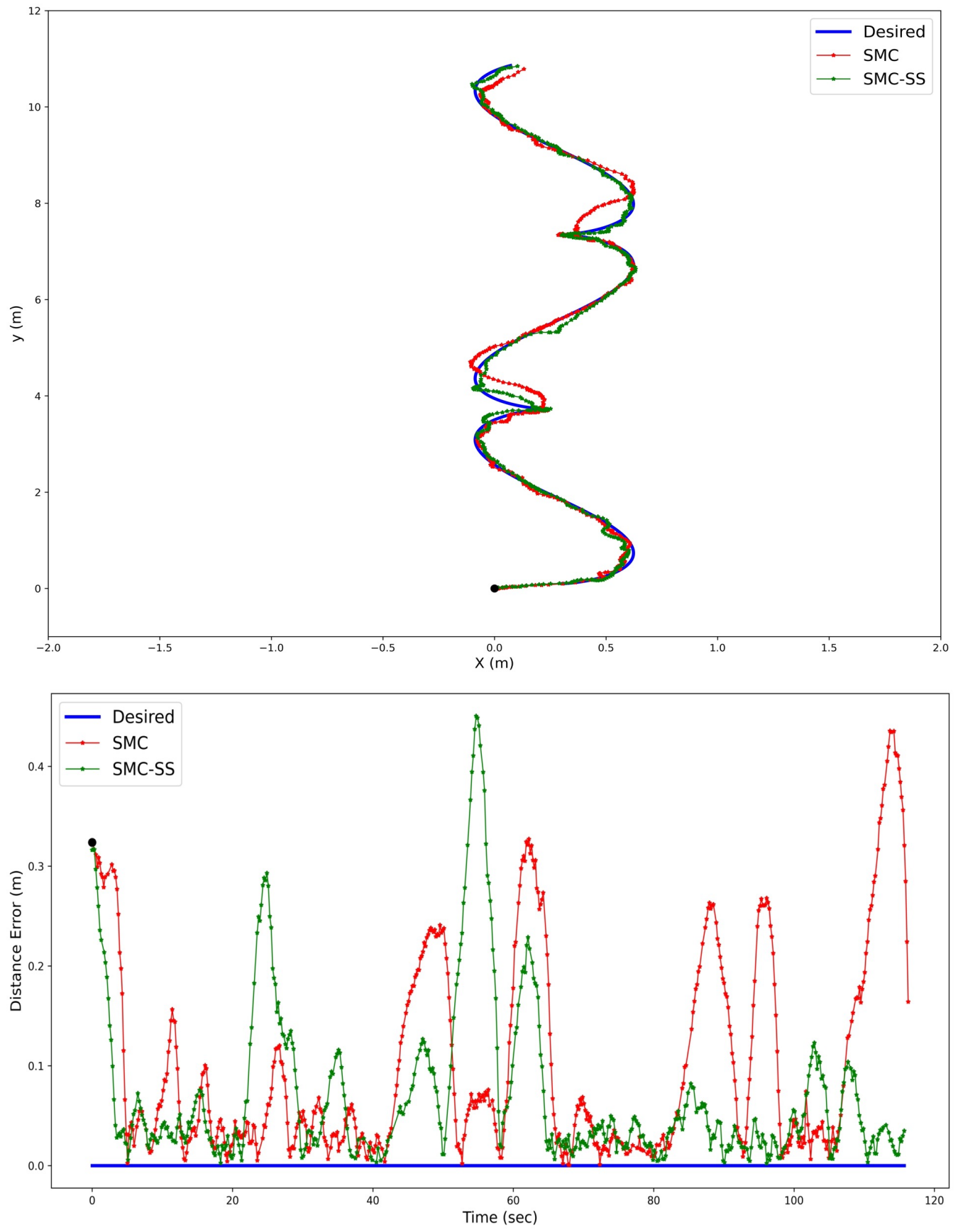

Figure 12. Experimental results of the SMC and SMC-SS controllers following the bow-shape trajectory; (top) trajectory, (bottom) distance error.

Figure 13 and Figure 14-bottom show the robot's heading angle and $e_\theta$. The desired heading angle has two cusp points around 40 and 80 seconds, where the robot needs to harshly change its direction to be able to follow the desired trajectory. These figures and Table 7 show that the SMC-SS controller performed better than the other controller by 10.92% and 9.89% in the mean of $|e_\theta|$ RMS of $e_\theta$, respectively. Overall, results show that the SMC-SS controller provides a more accurate performance in tracking the bow-shape trajectory due to its compensation for vehicle slipping and skidding.

*D. Significance Analysis*

Table 6 and Table 7 summarize the performance of the SMC and SMC-SS controllers across all experiments for all trajectories. Overall, the experimental results in Table 7 show better performance of the sliding-mode controller with slip and undesired skid compensation at the vehicle-level. This table shows, on average, more than 27% improvement in the distance error for the three manoeuvres. This table indicates that the highest and lowest improvement of the mean of distance errors are achieved for the circular and bow-shape trajectories by about 38% and 27%, respectively. The circular trajectory is a challenging path for the SSMR robots and the slip and undesired skid compensation reduced the tracking error. On the other hand, the bow-shape trajectory is challenging for the SSMR as well as the compensation system. During the bow-shape trajectory, the robot experiences two sharp rotations with low speeds at cusp points, which decreases the SNR and causes difficulty for the estimators. However, the SMC-SS still improved the performance of tracking the bow-shape trajectory similar to the straight-line one.

Table 6. The average performance of the SMC controller tracking the three desired trajectories under uneven grass terrain conditions. M: mean of, *dis* in cm, and $e_\theta$ in degree

| **Manoeuvre** | **SMC** | | | |
|---|---|---|---|---|
| | **M *dis*** | **RMS *dis*** | **M $\|e_\theta\|$** | **RMS $e_\theta$** |
| **Straight-line** | 15.55 | 26.89 | 3.65 | 4.71 |
| **Circular** | 12.24 | 19.33 | 13.65 | 14.73 |
| **Bow-shape** | 9.74 | 13.81 | 20.79 | 35.70 |

Table 7. The average performance of the SMC-SS controller tracking the three desired trajectories under uneven grass terrain conditions. M: mean of, *dis* in cm, and $e_\theta$ in degree. (%) indicates percentage improvement in comparison with the SMC controller.

| **Manoeuvre** | **SMC-SS** | | | |
|---|---|---|---|---|
| | **M *dis*, (%)** | **RMS *dis*, (%)** | **M $\|e_\theta\|$, (%)** | **RMS $e_\theta$, (%)** |
| **Straight-line** | 11.21, +27.91 | 21.03, +21.79 | 1.84, +49.59 | 2.47, +47.56 |
| **Circular** | 7.55, +38.32 | 12.76, +33.99 | 12.08, +11.50 | 13.34, +9.44 |
| **Bow-shape** | 7.07, +27.41 | 10.17, +26.36 | 18.52, +10.92 | 32.17, +9.89 |

For further comparison, the FAR test was applied to the distance error across the three trajectories and the result is shown in Table 8. The FAR test result between SMC and SMC-SS is 0.034, which means the null hypothesis of the test is rejected and it can be concluded that the controllers do not perform similarly. Therefore, the post hoc Finner test was utilized and the result indicates that there is a significant difference between the two designed controllers.

Table 8. Significance test results between SMC and SMC-SS controllers. R: rejected

| **Controller** | **FAR** | **Post hoc Finner test** |
|---|---|---|
| **SMC vs SMC-SS** | 0.034 (R) | 0.004 |

*E. Slip and Undesired Skid Estimator*

During this experimental study, the previously developed slip and undesired skid estimators [6], [8] were utilized to design the control-feedback loop and include compensation for these effects in outdoor environments.

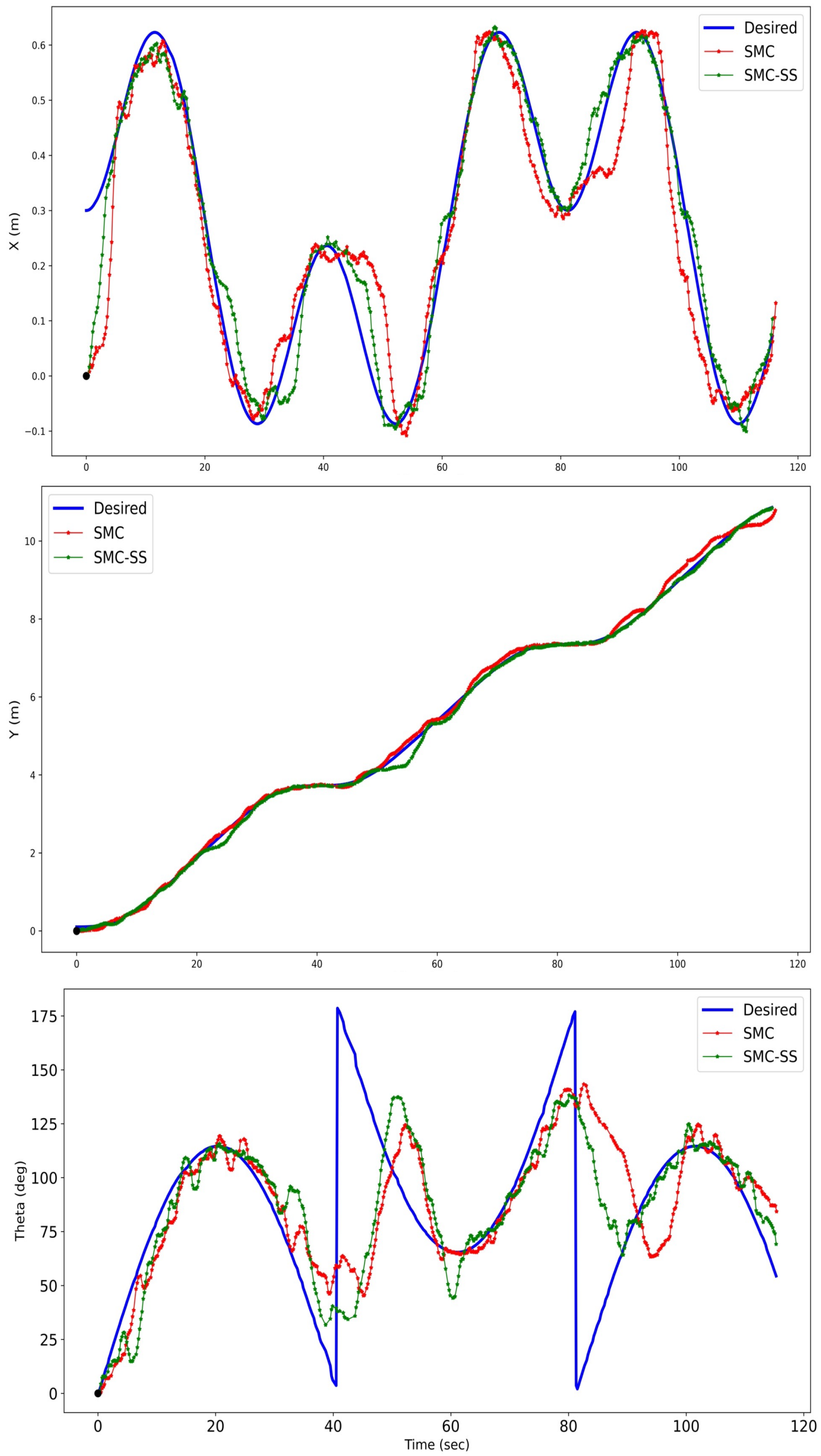

Figure 13. Trajectories of the SMC and SMC-SS controllers following the bow-shape trajectory; (top) $x$, (middle) $y$, (bottom) $\theta$.

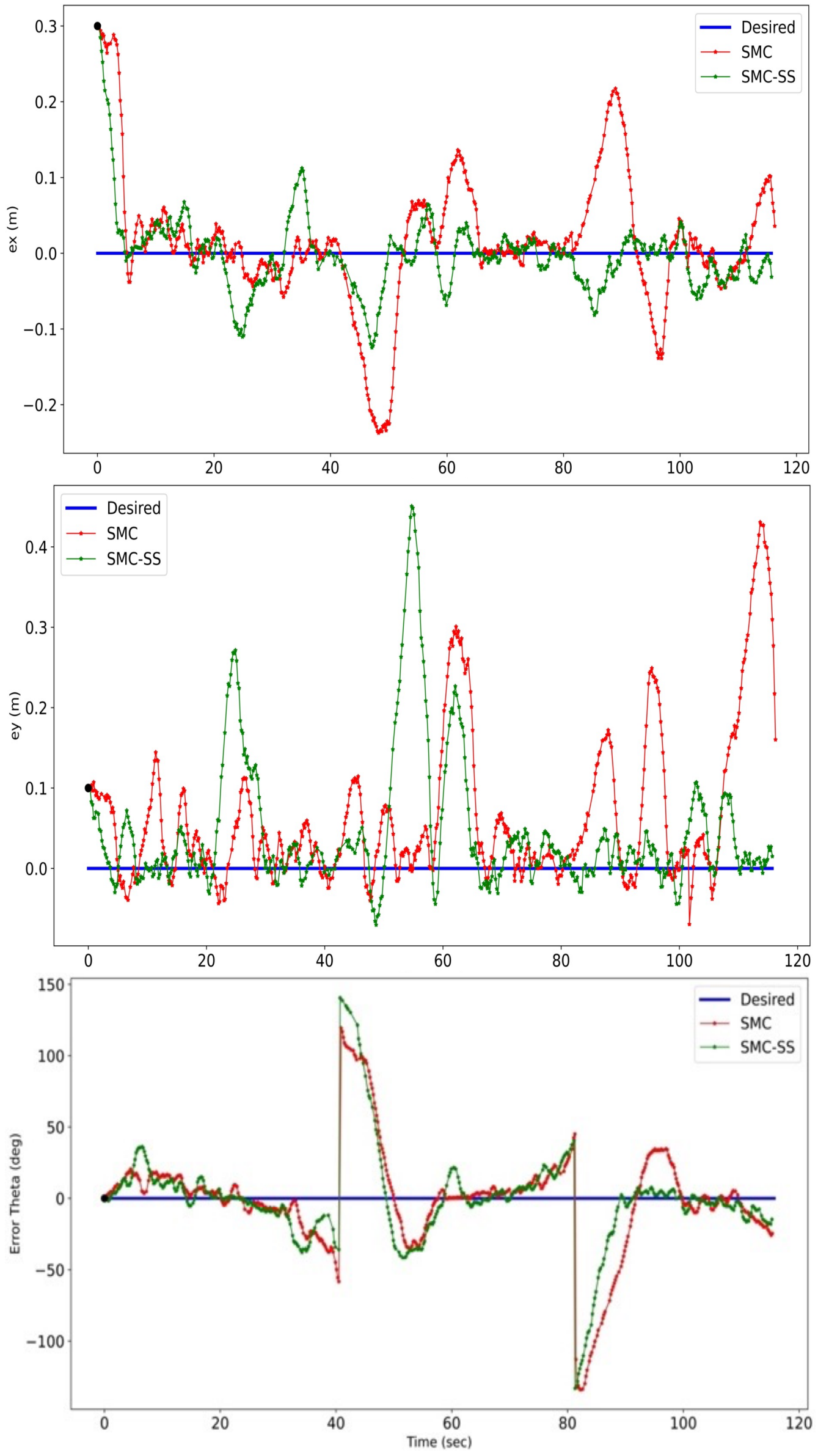

Figure 14. Tracking errors of the SMC and SMC-SS controllers following the bow-shape trajectory; (top) $e_x$, (middle) $e_y$, (bottom) $e_\theta$.

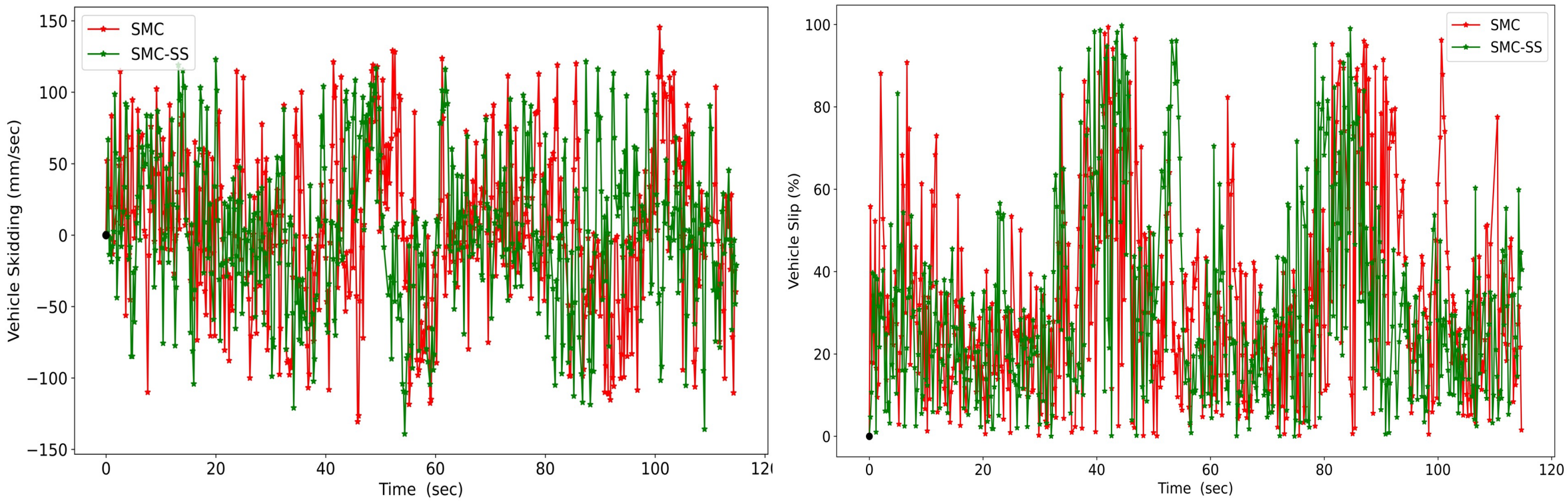

Figure 15. The robot's actual undesired skidding (left) and slipping (right) following the bow-shape trajectory.

Although both CNN-LSTM-AE and CNN-LSTM models were tested before [6], [8], the performance of these models was again evaluated during the trajectory tracking experiments of this paper. Note that the performance of the slip and undesired skid estimators were evaluated using all 8 manoeuvres' data of each controller (e.g., straight-line: 3, circular: 3, and bow-shape: 2 experiments), but only the first experiment of each trajectory is visualized here.

Table 9 and Table 10 show the performance of the slip and undesired skid estimators on grass, respectively. These tables show that the slip and undesired skid estimators respectively perform with 7.06%, 15.74% and 12.35 *mm/sec*, 26.43% in MAE and SMAPE.

The slip estimator's performance in this paper's experiments slightly differs from our previous slip estimation paper [8]. The first reason for this difference could be because of testing the robot only on grass for these experiments. Whereas the robot was driven on grass, sand, clay, and gravel for the performance analysis in the slip estimation paper [8]. The second reason could be because of driving the robot on the bow-shape trajectory, where the slip estimator deals with higher estimator errors due to having a low SNR at cusp points. It is worth mentioning that the slip estimator performs with a 5.14% MAE score when only straight-line and circular trajectories are considered, which is similar to the performance of the slip estimator in slip estimation paper [8]. Moreover, due to having high initial errors, the robot starts with the maximum wheels' rotation speeds at the beginning of all manoeuvres, and as a result, high estimation errors at those moments [8].

In contrast, the undesired skid estimator performs almost similarly to the performance of the model in our skid estimation paper [6]. The reason could be because of the robustness of the velocity-based undesired skid definition to the measurement noises, which helped the model to not get affected by the low SNR during the manoeuvres, especially at cusp points and the beginning moment of the robot's movement.

Table 9. Regression and classification results of the slip estimator during the 8 trajectory tracking experiments (↑, ↓: highest and lowest number desirable, respectively).

| Model | Regression | | Classification |
|---|---|---|---|
| | MAE (↓) | SMAPE (↓) | F1 (↑) |
| **CNN-LSTM-AE** | 7.06 | 15.74 | 87.01 |

Table 10. Regression and classification results of the undesired skid estimator during the 8 trajectory tracking experiments (↑, ↓: highest and lowest number desirable, respectively).

| Model | Regression | | Classification |
|---|---|---|---|
| | MAE (↓) | SMAPE (↓) | Acc (↑) |
| **CNN-LSTM** | 12.35 | 26.43 | 97.73 |

Figure 16 shows the response of the slip estimator during the first experiment of each trajectory. In this figure, the first 900 samples belong to the straight-line and circular trajectories and the remainder represents the robot on the bow-shape trajectory. Figure 16-top and middle show higher errors at four cusps of the bow-shape trajectory (e.g., two for the SMC and two for the SMC-SS controller) at 1100, 1300, 1650, and 1850 samples. Figure 16-bottom shows the distribution of the estimation error. This distribution, similar to the distribution of the estimation error presented in slip estimation paper [8], shows that the residuals are roughly evenly distributed around zero, which is desirable.

Figure 17 shows the performance of the undesired skid estimator during the first experiment of each trajectory. This figure shows that the CNN-LSTM model not only performs consistently well throughout all three trajectories but also follows the right direction of the actual undesired skidding with 97.73% accuracy (Table 10).

## V. Conclusion

This paper presented a trajectory-tracking controller with slip and undesired skid compensation at the vehicle-level using sliding-mode control and deep learning techniques for outdoor environments. The kinematics model of the SSMR was modified to consider the slipping and undesired skidding at the vehicle-level, and sliding-mode controller was utilized to regulate the tracking errors. The robot's slipping and undesired skidding were estimated using the two previously validated deep learning models in our other works [6], [8] and then fed to the control-feedback loop to compensate for them. Three desired trajectories were defined for the robot to follow on uneven grass terrain. The results showed that the proposed slip and undesired skid compensation technique improved the mean of tracking distance error by more than 27%, highlighting the efficacy of the developed compensation technique in real-world scenarios. It is essential to note that while the evaluation focuses on SSMRs, the proposed slip and undesired skid compensation technique is not limited to this specific type of mobile robot, showcasing its potential applicability across various robotic platforms.

The novel contribution lies in the integration of robust control and deep learning techniques, enabling real-time compensation for slip and undesired skid at the vehicle-level operating in unforeseen outdoor terrains. By redefining the WTI representation using just two slip and undesired skid parameters at the vehicle-level, rather than the conventional approach requiring two slip parameters for each wheel, this research remarkably simplifies the compensation process. This paper shows the capacity of the previously developed deep learning slip and undesired skid estimators in a real-time implementation to significantly improve the performance of the trajectory tracking system in outdoor environments. This work not only refines our understanding of the effect of WTI on the dynamics of the robot but also paves the way for more streamlined and effective navigation strategies in real-world, unpredictable outdoor environments.

## Acknowledgment

The authors would like to thank Dr. Peter Donelan for his valuable comments on the singularity problem of the sliding-mode controller.

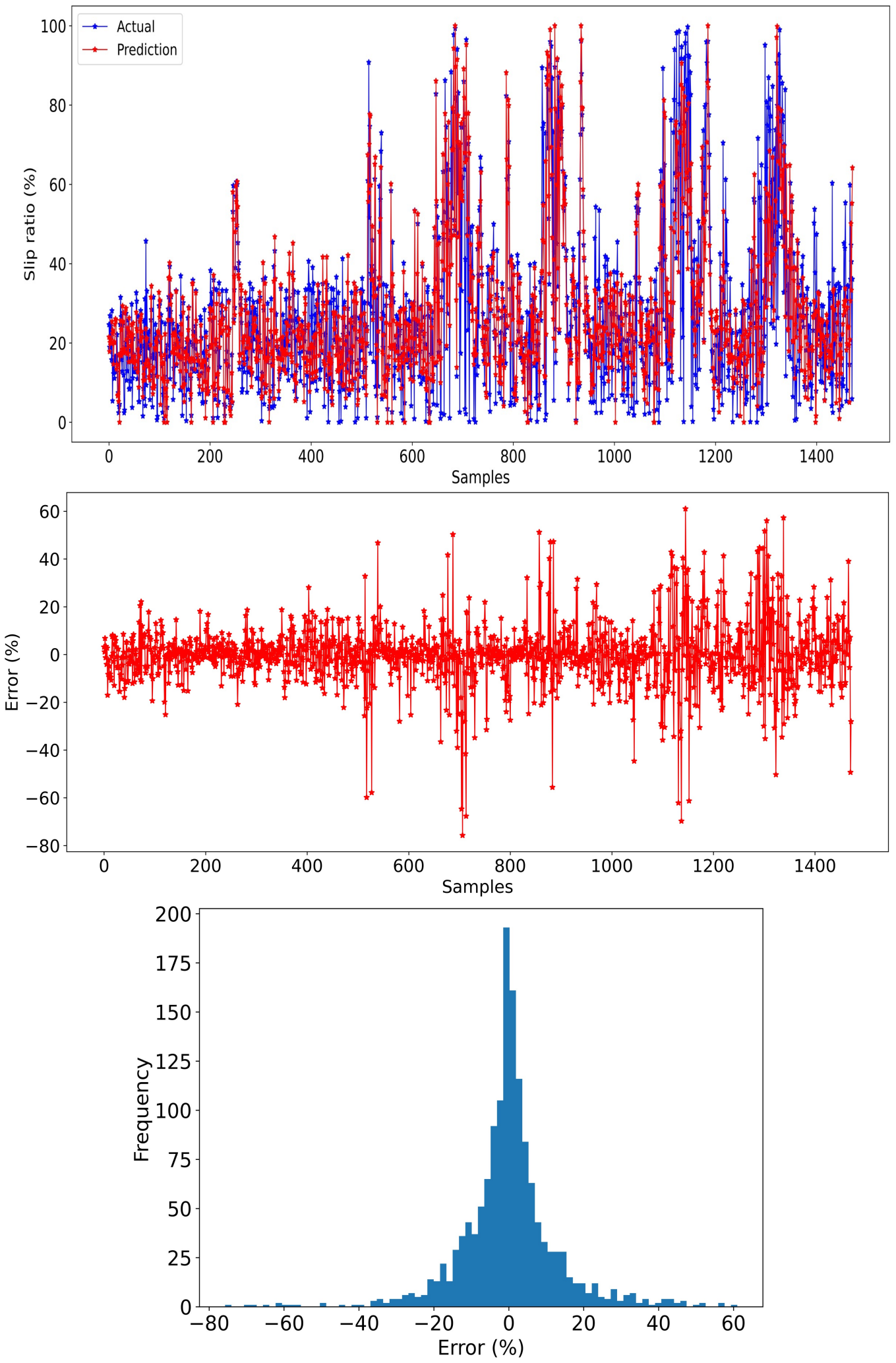

Figure 16. Performance of the slip estimator during the trajectory experiments; Top: actual slip ratio versus prediction, middle: estimation error, bottom: distribution of estimation error.

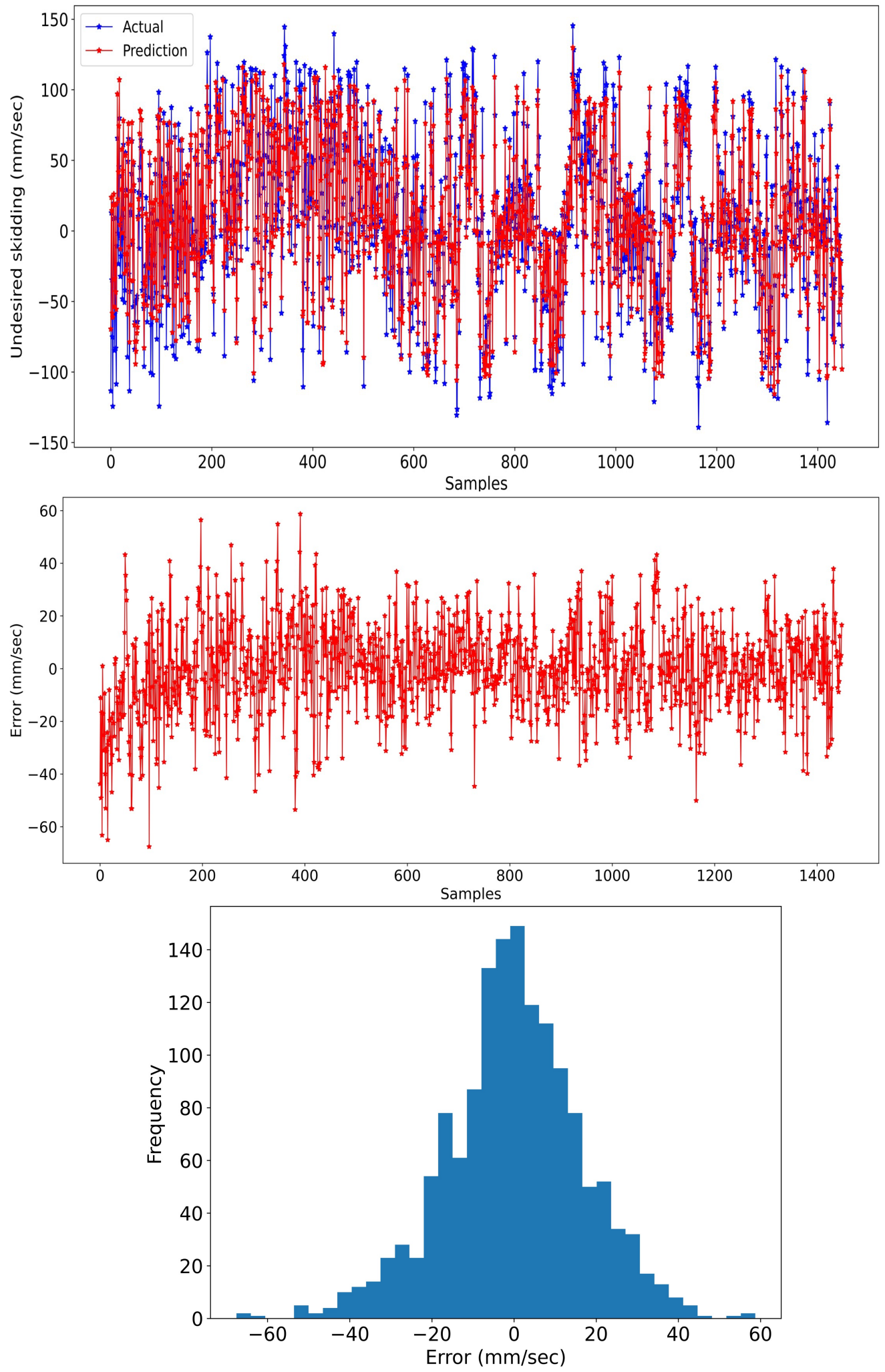

Figure 17. Performance of the undesired skid estimator during the trajectory experiments; Top: actual undesired skid versus prediction, middle: estimation error, bottom: distribution of estimation error.